\pdfoutput=1
\pdfoutput=1
\pdfoutput=1
\pdfoutput=1
\pdfoutput=1
\documentclass[10pt,twocolumn,letterpaper]{article}

\usepackage{iccv}
\usepackage{times}
\usepackage{epsfig}
\usepackage{graphicx}
\usepackage{amsmath}
\usepackage{amssymb}
\usepackage{graphicx}
\usepackage{amsmath}
\usepackage{amssymb}
\usepackage{booktabs}
\usepackage{bbding}
\usepackage{svg}
\usepackage{bm}
\usepackage{ulem}
\usepackage{multirow}
\usepackage{float}
\usepackage[breaklinks=true,bookmarks=false]{hyperref}

\iccvfinalcopy 

\def\iccvPaperID{2459} 

\ificcvfinal\fi

\begin{document}

\title{ MXM-CLR: A Unified Framework for Contrastive Learning of\\ Multifold Cross-Modal Representations}

\author{Ye Wang$^1$ \hspace{3em}
Bowei Jiang$^1$  \hspace{3em}
Changqing Zou$^{2,3}$ \hspace{3em}
Rui Ma$^{1,4}$\thanks{Rui Ma is the corresponding author.}
\\
$^1$Jilin University \hspace{3em} $^2$Zhejiang University \hspace{3em} $^3$Zhejiang Lab \\
$^4$Engineering Research Center of Knowledge-Driven Human-Machine Intelligence, MOE\\
{\tt\small \{yewang22,jiangbw22\}@mails.jlu.edu.cn \hspace{1em} aaronzou1125@gmail.com \hspace{1em} ruim@jlu.edu.cn}
}


\maketitle
\ificcvfinal\thispagestyle{empty}\fi

\definecolor{green}{rgb}{0, 0.5, 0}
\definecolor{orange}{rgb}{0.8, 0.6, 0.2}
\definecolor{red}{rgb}{1.0, 0.0, 0.0}
\definecolor{teal}{rgb}{0.0, 0.4, 0.4}
\definecolor{purple}{rgb}{0.65,0,0.65}
\definecolor{saffron}{rgb}{0.95,0.75,0.2}
\definecolor{turquoise}{rgb}{0.0,0.5,0.5}
\definecolor{black}{rgb}{0.0, 0.0, 0.0}
\definecolor{gray}{rgb}{0.5, 0.5, 0.5}

\newcommand{\rui}[1]{{\color{purple}#1}}
\newcommand{\wy}[1]{{\color{black}#1}}
\newcommand{\TODO}[1]{{\color{red}#1}}

\begin{abstract}
Multifold observations are common for different data modalities, e.g., a 3D shape can be represented by multi-view images and an image can be described with different captions.
Existing cross-modal contrastive representation learning (XM-CLR) methods such as CLIP are not fully suitable for multifold data as they only consider one positive pair and treat other pairs as negative when computing the contrastive loss.
In this paper, we propose MXM-CLR, a unified framework for contrastive learning of multifold cross-modal representations.
MXM-CLR explicitly models and learns the relationships between multifold observations of  instances from different modalities for more comprehensive representation learning.
The key of MXM-CLR is a novel multifold-aware hybrid loss which considers multiple positive observations when computing the hard and soft relationships for the cross-modal data pairs.
We conduct quantitative and qualitative comparisons with SOTA baselines for cross-modal retrieval tasks on the Text2Shape and Flickr30K datasets.
We also perform extensive evaluations on the adaptability and generalizability of MXM-CLR, as well as ablation studies on the loss design and effects of batch sizes.
The results show the superiority of MXM-CLR in learning better representations for the multifold data. The code is available at \url{https://github.com/JLU-ICL/MXM-CLR}.
\end{abstract}

\section{Introduction}

\begin{figure}
  \centering
  \includegraphics[width=0.9\linewidth]{teaser.pdf}
    \caption{Illustration of MXM-CLR as a unified framework for contrastive learning of multifold cross-modal representations.
    Different XM-CLR models such as CLIP and TriCoLo can be derived from MXM-CLR by adjusting the data pairing and the relationship modeling schemes. 
    The gray dashed boxes represent training batches with multiple positive observations. 
    The solid or dash lines indicate the hard or soft relationships between the cross-modal data pairs.
    Note the dashed border of T3 in (d) shows the observation may not be sampled into the batch. 
    }
     \vspace{-10pt}
  \label{fig:teaser}
\end{figure}
\label{sec:intro}

Representation learning is a key task for deep learning \cite{YoshuaBengio2013RepresentationLA}.
Extensive efforts have been paid on learning robust and generalizable feature representations for data of different modalities, such as text \cite{JacobDevlin2018BERTPO,AshishVaswani2017AttentionIA}, images \cite{KaimingHe2015DeepRL,AlexeyDosovitskiy2020AnII} and point clouds \cite{Qi_2017_CVPR}. 
Recently, cross-modal representation learning that aims to learn a joint embedding space of features representing two different data modalities has attracted wide attention for its emerging applications in vision-language pretraining \cite{AlecRadford2021LearningTV,wang2022image,JunkeWang2022OmniVLOneFM,YanZeng2021MultiGrainedVL}, text-to-image generation \cite{AdityaRamesh2021ZeroShotTG,ramesh2022hierarchical,nuwa,Saharia2022PhotorealisticTD}, text-based 3D retrieval \cite{chen2018text2shape,ZhizhongHan2019Y2Seq2SeqCR,ChuanTang2021Part2WordLJ,YueRuan2022TriCoLoTC} and 3D visual grounding \cite{PanosAchlioptas2020ReferIt3DNL,DaveZhenyuChen2020ScanRefer3O,PinHaoHuang2021TextGuidedGN,JunhaRoh2021LanguageReferSM,ZhihaoYuan2021InstanceReferCH,LichenZhao20213DVGTransformerRM} etc.

Contrastive learning is a type of representation learning scheme that can effectively learn discriminative features in a supervised \cite{PrannayKhosla2022SupervisedCL} or unsupervised manner \cite{ZhirongWu2018UnsupervisedFL,TingChen2020ASF}.
The key idea is to construct two kinds of data pairs, i.e., positive/negative pairs corresponding to the same/different instances, and design losses so that the learned features for the positive pairs are pulled together, while the features for negative pairs are repelled from each other.
For example, InfoNCE \cite{AaronvandenOord2018RepresentationLW,YuhaoZhang2021ContrastiveLO} is a widely-used contrastive loss for representation learning and it is designed to maximally preserve the mutual information of the positive pairs.

Naturally, contrastive learning can be applied to learning inherent features for data coming from the same modality \cite{ZhirongWu2018UnsupervisedFL,AaronvandenOord2018RepresentationLW,YonglongTian2019ContrastiveMC,KaimingHe2022MomentumCF,XinleiChen2020ImprovedBW,TingChen2020ASF,TingChen2020BigSM,PrannayKhosla2022SupervisedCL} as well as to aligning features for cross-modal data \cite{AlecRadford2021LearningTV,ChaoJia2021ScalingUV,JunnanLi2021AlignBF,yang2022vision,YueRuan2022TriCoLoTC}.
In the pioneering work CLIP \cite{AlecRadford2021LearningTV}, cross-modal contrastive learning (XM-CLR) is performed on 400 million image-text pairs collected from the Internet to learn transferable language and image representations.
Beyond text and images, XM-CLR has also been explored on 3D data \cite{YueRuan2022TriCoLoTC,zhang2022pointclip,sanghi2022clip,hong2022avatarclip}.
For example, TriCoLo \cite{YueRuan2022TriCoLoTC} employs constrastive learning to train a joint embedding space and obtain aligned representations of 3D colored voxels, text descriptions and multi-view images for shapes in the Text2Shape dataset \cite{chen2018text2shape}.

Although existing XM-CLR methods have achieved intriguing results on learning aligned features for different data modalities, they are mainly single-fold-oriented as only one positive pair is considered when computing the contrastive loss.
On the other hand, the representation of a particular instance is often \textit{multifold}.
For example, as shown in Figure \ref{fig:teaser}(a), multiple text sentences can be used to describe a 3D table and the table can also be rendered into images at different views. 
Hence, the observations of a particular table are multifold for either the text or image modality.
Such multifold observations are also common for text descriptions of natural images, e.g., each image in Flickr30K \cite{PeterYoung2014FromID} and MS COCO \cite{lin2014microsoft} is annotated with about five captions describing the fine-grained content of the image.
With the multifold cross-modal observations, it is easy to obtain multiple positive pairs using the Cartesian product operation.
These additionally constructed positive pairs contain useful information for learning the representation of the corresponding data instance.
However, how to properly utilize these positive pairs in a contrastive learning framework has not been explored before.

In this paper, we propose \textbf{MXM}-CLR, a unified framework for \textbf{m}ultifold \textbf{c}ross-\textbf{m}odal contrastive representation learning.
Our key insight is, for cross-modal constrastive learning on data with multifold observations, multiple positive pairs could be conveniently constructed and the loss function should be designed to simultaneously leverage these positive pairs for more comprehensive feature learning.
Thus, we propose a novel multifold-aware hybrid loss, named MFH, to consider multiple positive observations when computing the hard and soft relationships for the cross-modal data pairs.
Specifically, we first construct \textit{group-wise} data pairings by applying Cartesian product to the observations of each modality.
In this sense, the MXM in the MXM-CLR acronym can also be understood as the Cartesian product between multifold observations of two modalities.
Then, the MFH models the hard relationships between multiple observations based on a subset of positive pairs obtained with a group-wise random selection scheme.
In addition, the soft relationships between the observations in both positive and negative pairs are learned based on the pseudo-targets generated by a momentum model \cite{li2021align}.
As more information from multiple observations is utilized by the MFH loss in a hybrid manner, our MXM-CLR can achieve superior performance on representation learning.

Comparing to previous single-fold-oriented methods such as CLIP, MXM-CLR is primarily suitable for data with multifold observations.
Meanwhile, as a unified framework, MXM-CLR can also be derived to different forms, while each variation corresponds to a model for a particular cross-modal task.
For example, as shown in Figure \ref{fig:teaser}(a), CLIP is a special case of MXM-CLR that only considers the hard relationships based on three given positive pairs in a one-to-one mapping manner.
Also, as shown in Figure \ref{fig:teaser}(d), TriCoLo (I), which is the bimodal version of TriCoLo \cite{YueRuan2022TriCoLoTC}, can be regarded as a degenerate case of MXM-CLR that is trained with multifold text features and the aggregated multi-view image feature.
Furthermore, MXM-CLR can be adapted to model other contrastive learning scenarios, including learning from the aggregated text feature and multifold image features (Figure \ref{fig:teaser}(e)), and from the aggregation of features for both modalities (Figure \ref{fig:teaser}(f)).
These two variations may correspond to novel tasks such as comprehensive text-to-image and text-to-shape retrieval, respectively.

To evaluate the performance of MXM-CLR, we perform extensive experiments on different cross-modal retrieval tasks on Text2Shape \cite{chen2018text2shape} and Flickr30K \cite{PeterYoung2014FromID}.
From the quantitative and qualitative results,
MXM-CLR models trained with MFH loss outperform the SOTA methods and multiple strong baselines with a significant margin.
We also perform ablation studies to validate the effect of each term of the MFH loss
and evaluate how the batch size can affect the results for MXM-CLR. 

In summary, our main contributions are as follows:

\begin{itemize}
    \item We propose MXM-CLR, the first unified framework for contrastive representation learning of cross-modal data with multifold observations, and we show existing XM-CLR methods such as CLIP \cite{AlecRadford2021LearningTV} and TriCoLo \cite{YueRuan2022TriCoLoTC} can be regarded as special cases of MXM-CLR.
    
    \item We introduce MFH, a novel multifold-aware hybrid loss, to explicitly model the hard and soft relationships for the data batch constructed with multiple positive observations.
    
    \item Extensive quantitative and qualitative comparisons with SOTA methods and strong baselines show the superiority of MXM-CLR in learning better representations for multifold data. 
    
\end{itemize}

\section{Related Work}
\label{sec:related_work}

\textbf{Contrastive representation learning.}
The main idea of contrast learning is to pull data points in positive pairs together and push data in negative pairs away.
Due to its simplicity and flexibility, contrastive learning has attracted numerous interests for representation learning \cite{ZhirongWu2018UnsupervisedFL,AaronvandenOord2018RepresentationLW,YonglongTian2019ContrastiveMC,KaimingHe2022MomentumCF,XinleiChen2020ImprovedBW,TingChen2020ASF,TingChen2020BigSM,PrannayKhosla2022SupervisedCL}.
Wu et al. \cite{ZhirongWu2018UnsupervisedFL} first propose a contrastive loss (named NCE loss) for the instance discrimination task.
Based on NCE loss, CPC \cite{AaronvandenOord2018RepresentationLW} introduces InfoNCE loss and  utilizes the autoregressive model to learn data representations by predicting the future in latent space. 
Comparing to above works which mainly deal with data in a single modality, we focus on cross-modal contrastive learning and propose to learn the joint embeddings of data with multifold observations.
Our multifold-aware MFH loss can simultaneously leverage multiple positive observations for more comprehensive feature learning.

\textbf{Joint learning of cross-modal data.}
A visual concept can be described with different data modalities including vision, language and 3D etc.
A great deal of efforts have been paid to the joint learning of vision and language features \cite{AlecRadford2021LearningTV,ChaoJia2021ScalingUV,JunnanLi2021AlignBF,yao2021filip,yang2022vision,wang2022image,FartashFaghri2018VSEIV,AndreaFrome2013DeViSEAD,RyanKiros2014UnifyingVE,YuhaoZhang2021ContrastiveLO,JunkeWang2022OmniVLOneFM}.
Among these works, CLIP \cite{AlecRadford2021LearningTV} is a pioneering work that is trained using InfoNCE to learn transferable features between image and text.
In addition, joint learning of 3D-image \cite{li2015joint,uy2021joint,qi2021toward,liu2021fine,zhu2019label,fu2020hard} and 3D-text \cite{chen2018text2shape,ZhizhongHan2019Y2Seq2SeqCR,ChuanTang2021Part2WordLJ,YueRuan2022TriCoLoTC}, as well as the 3D-involved applications such as image-based 3D shape retrieval \cite{lin2021single,kuo2021patch2cad,fu2020hard}, image based 3D reconstruction \cite{peng2022tmvnet,zhang2021view}, 3D visual grounding \cite{PanosAchlioptas2020ReferIt3DNL,DaveZhenyuChen2020ScanRefer3O} and text-to-3D generation \cite{sanghi2022clip,hong2022avatarclip}, have also attracts great attention in graphics and vision community.
Comparing to above methods, we are the first to propose a unified cross-modal joint learning framework which can be easily adapted to different tasks and different data modalities.

\textbf{Relationship modeling with different hard pairings.}
To improve the performance for representation learning, several works have been proposed to model more complicated hard relationships of the data samples.
In contrast to CLIP which only considers pairwise positive or negative relationships in a one-to-one manner, CMC \cite{YonglongTian2019ContrastiveMC} treats all samples from different sensors as positive to maximize mutual information between different views (modalities) of the same scene.
SupCon \cite{PrannayKhosla2022SupervisedCL} leverages the label information and treats all samples with the same class label as positive.
Instead of modeling relationships for the positive pairs, CrossCLR \cite{MohammadrezaZolfaghari2021CrossCLRCC} identifies the highly related text-video samples and masks them from negative samples to avoid issues with false negatives for InfoNCE-based methods.
Comparing to these methods, we construct groups of positive pairs from different observations of the same instance and apply a group-wise masked random selection scheme to obtain multiple combinations of different positive pairs when performing the contrastive learning.

\textbf{Relationship modeling with soft-target learning.}
Apart from modeling the hard relationships for the data pairs, a few works \cite{li2021align,cheng2021data,andonian2022robust,zhong2022regionclip} take the idea of the knowledge distillation and employ a teacher model to generate the soft-alignment targets for learning the relationships between the paired data.
For example, ALBEF \cite{li2021align} uses a momentum model as a continuously-evolving teacher to generate the pseudo-targets as additional supervision for training.
Similarly, Robust-XR \cite{andonian2022robust} adopts progressive self-distillation to generate soft image-text alignments for robust representation learning from noisy data.
Also, HCMoCo attempts \cite{hong2022versatile} to explore the intra- and inter-sample hard and soft relationships for human-centric representation learning.
For our MXM-CLR, we aim to mine more comprehensive relationships from the multifold observations, while such setting can bring both benefits and challenges for the soft relationship learning.
Hence, we propose a multifold-aware hybrid learning strategy by combining the soft-target learning with the group-wise masked hard relationship modeling to boost the learning performance.

\section{Method}
\label{method}

\begin{figure}
  \centering
  \includegraphics[width=0.95\linewidth]{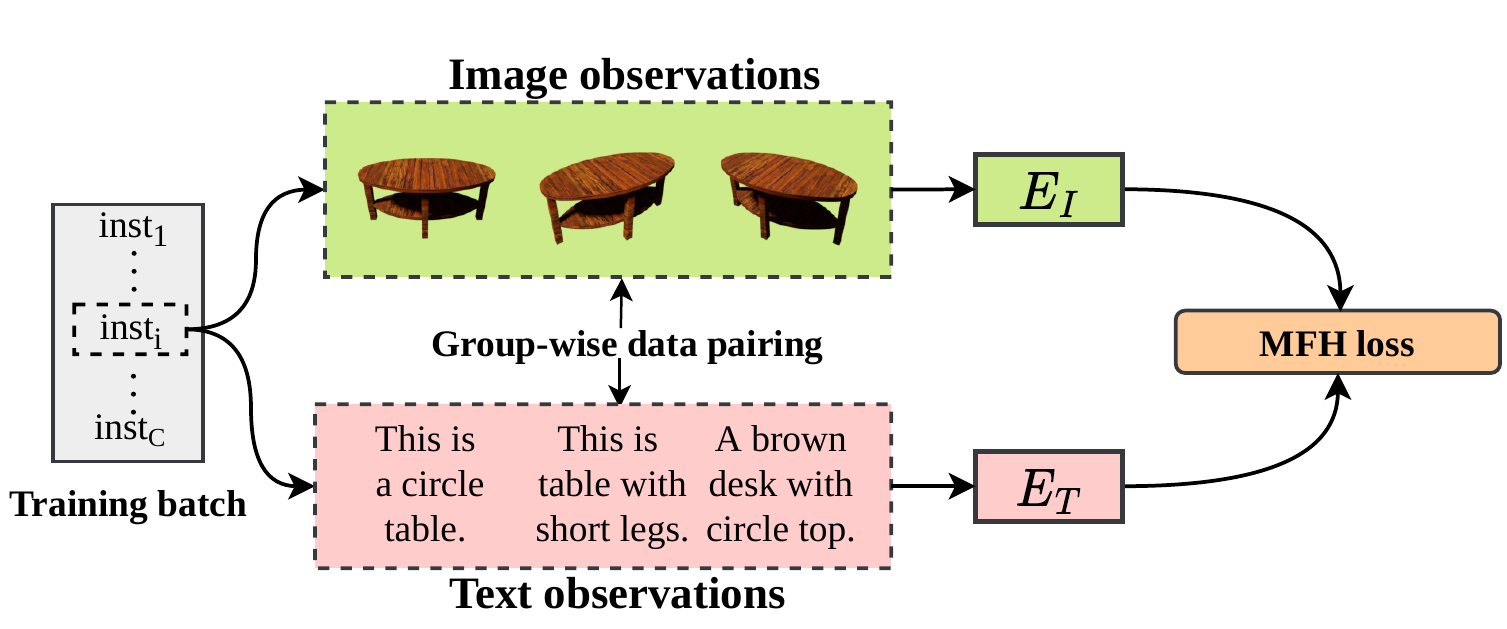}
   \caption{ Overview of MXM-CLR framework. 
   }
   \vspace{-8pt}
   \label{fig:framework}
\end{figure}

The overview of MXM-CLR framework is shown in Figure \ref{fig:framework}.
For each instance in the mini-batch, the cross-modal pairs are constructed from its multifold text and image observations following the group-wise data pairing process, while the feature of each observation is extracted using the corresponding encoder.
Then, the MFH loss is calculated to model hard and soft relationships for the cross-modal observations.
In the following, we first revisit the representative XM-CLR methods such as CLIP \cite{AlecRadford2021LearningTV} and TriCoLo \cite{YueRuan2022TriCoLoTC}, as well as the InfoNCE \cite{AaronvandenOord2018RepresentationLW,YuhaoZhang2021ContrastiveLO} loss adopted by these methods.
Next, we introduce the MXM-CLR framework and present details for group-wise data pairing and the MFH loss.

\begin{figure*}
  \centering
  \includegraphics[width=0.82\linewidth]{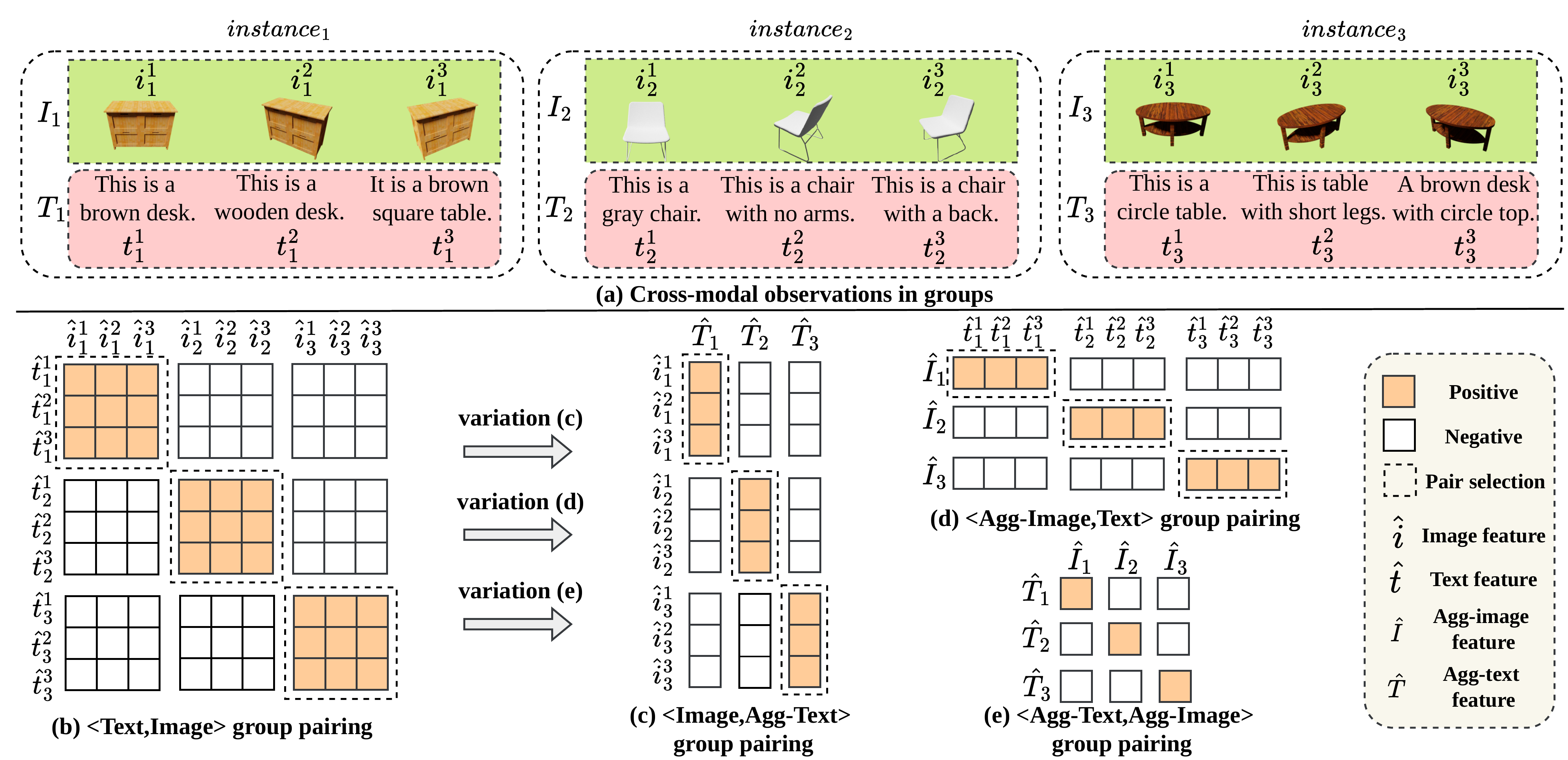}
  \vspace{-2pt}
  \caption{
  Given a mini-batch containing three instances (a), group-wise data pairing can be performed to construct positive and negative pairs for constrastive learning.
  Group-wise pairing results for different variations of MXM-CLR are shown in (b), (c), (d).
  The dashed boxes drawn around the positive groups indicate where the positive pair selection is performed. For (e), MXM-CLR degenerates to single-fold contrastive learning such as CLIP.
   }
   \vspace{-10pt}
   \label{fig:overview_loss}
\end{figure*}

\subsection{Revisiting XM-CLR}
\label{sec:xm-clr}
XM-CLR methods aim to learn a joint embedding space for cross-modal representation learning. CLIP \cite{AlecRadford2021LearningTV} and TriCoLo \cite{YueRuan2022TriCoLoTC} are two representative methods for representation learning of image-text and 3D-text, respectively.
For CLIP \cite{AlecRadford2021LearningTV}, it employs a dual-encoder architecture which consists of an image encoder (e.g., ResNet \cite{KaimingHe2015DeepRL} or ViT \cite{AlexeyDosovitskiy2020AnII}) and a text encoder (e.g., BERT \cite{JacobDevlin2018BERTPO}).
For TriCoLo \cite{YueRuan2022TriCoLoTC}, as the original version TriCoLo (I+V) learns from three modalities, i.e., text, image and voxel, we focus on its bimodal version TriCoLo (I) which also learns from text and image as ours.
Different from CLIP, TriCoLo aggregates the features from multi-view images by MVCNN \cite{HangSu2015MultiviewCN} and learns the alignments between text and the aggregated feature.
Based on the features obtained for each modality, CLIP and TriCoLo utilize InfoNCE loss \cite{AaronvandenOord2018RepresentationLW} to optimize the encoders for representation learning.

Given a mini-batch constructed from $N$ image-text pairs $\{(i^1,t^1),(i^2,t^2),\cdots,(i^N,t^N)$\}, we denote the features extracted by the encoders as $\{(\hat{i}^1,\hat{t}^1),(\hat{i}^2,\hat{t}^2),\cdots,(\hat{i}^N,\hat{t}^N)$\}.
For InfoNCE, it first defines the image-to-text loss as:
\begin{scriptsize}
\begin{equation}
\label{eq_1}
L_{I\rightarrow T}= \frac{-1}{N} \!\sum_{r=1}^{N}\log_{}{\frac{exp(s(\hat{i}^r,\hat{t}^r) )}{exp(s(\hat{i}^r,\hat{t}^r)) \!+\! \sum_{h{\in\! \mathcal{N}^r}}^{}\!\!exp(s(\hat{i}^r,\hat{t}^h))}},
\end{equation}
\end{scriptsize}%
where $s(\cdot,\cdot)$ is the temperature-controlled cosine similarity between features of cross-modal pairs;
$r$ and $h$ are indices for the samples;
$\mathcal{N}^r$ is the set of indices for negative samples of sample $r$.
The text-to-image loss $L_{T\rightarrow I}$ is defined symmetrically and the final InfoNCE loss is defined as:
\begin{equation}
L_{InfoNCE}=(L_{I\rightarrow T} + L_{T\rightarrow I})/2.
\end{equation}

\subsection{MXM-CLR Framework}
\label{sec:mxm}
In this section, we will introduce the architecture of MXM-CLR.
As MXM-CLR is a unified framework that can be applied to different modalities in various forms, we focus on the task of text-image representation learning while other forms of MXM-CLR such as text-shape learning can be derived similarly.

\textbf{Multifold cross-modal batch construction.}
For an image-text dataset $\mathcal{D}$ that contains $C$ instances, i.e., $\mathcal{D} = \{inst_1,inst_2,\cdots,inst_{C}\}$, we consider each instance has $m$ observations for image and $n$ observations for text: $inst_k = \{(i_k^1,i_k^2,\cdots,i_k^{m}),(t_k^1,t_k^2,\cdots,t_k^{n})\}$; see Figure \ref{fig:overview_loss}(a) for an example.
Then, we define the mini-batch for multifold data as:
\begin{equation}
\setlength{\abovedisplayskip}{3pt}
\setlength{\belowdisplayskip}{3pt}
mini\!\!-\!\!batch = \{inst_k | k\in\{1,2,\cdots,b\}\},
\end{equation}
where $b$ is the number of instances in a mini-batch.
Therefore, each mini-batch contains $b*m$ images and $b*n$ texts.

\textbf{Group-wise data pairing.}
Unlike the XM-CLR for which the positive and negative pairs can be directly obtained from the mini-batch, we need to generate positive and negative pairs based on the multifold observations for each data instance.
As different combinations of observations from each modality could form a positive pair, we employ Cartesian product to obtain all combinations of the cross-modal observations w.r.t. the same instance as positive pairs; then, the observations from different instances become negative pairs.
Figure \ref{fig:overview_loss}(b) shows an illustration of the resulting positive and negative pairs, and the positive pairs (in orange) naturally form groups in the diagonal.
Formally, the group consisting $m*n$ positive pairs of instance $k$ can be represented as:\par
\vspace{-10pt}
{\small
\begin{equation}
\setlength{\abovedisplayskip}{4pt}
\setlength{\belowdisplayskip}{4pt}
g_k = \{ (i_k^r,t_k^c)| r \in \{1,2,\cdots,m\},c \in \{1,2,\cdots,n\}\}.
\end{equation}
}%

\begin{figure}
  \centering
  \includegraphics[width=0.92\linewidth]{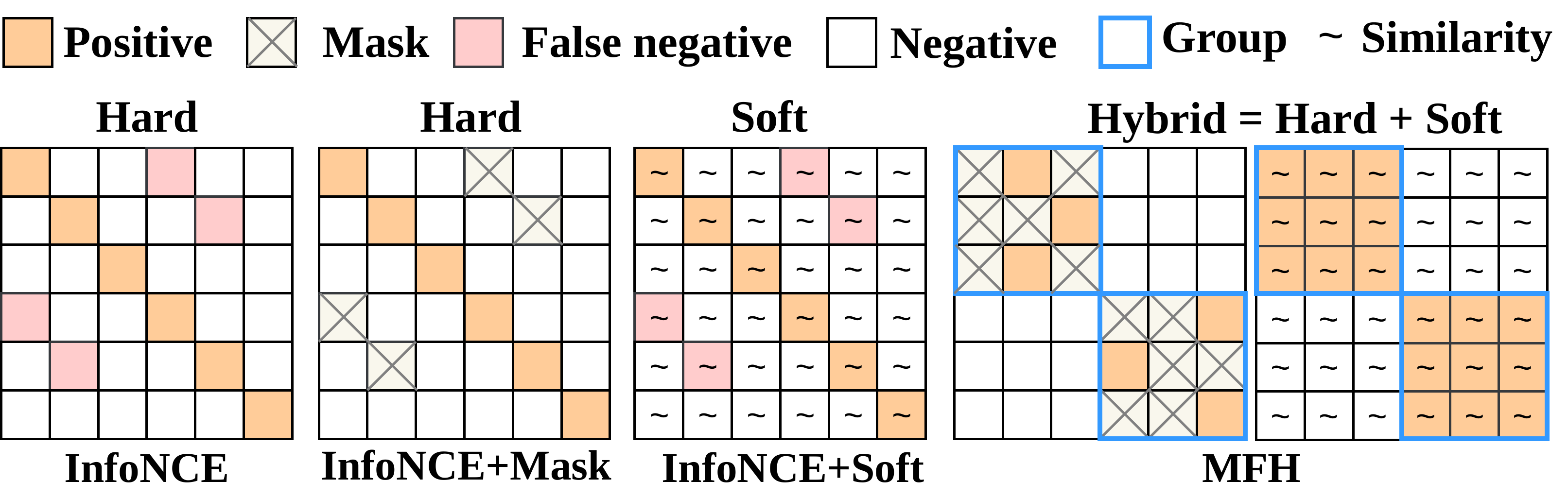}
   \caption{  
The illustration of how the positive and negative pairs are utilized to compute the data relationships in InfoNCE-based losses and the MFH loss.
   }
   \label{fig:infonce_mask_mf}
   \vspace{-8pt}
\end{figure}

\textbf{MFH loss.}
The core of MXM-CLR is the multifold-aware hybrid loss MFH, which explicitly models the hard and soft relationships for data with multiple observations.
Figure \ref{fig:infonce_mask_mf} shows how the MFH loss as well as some InfoNCE-based losses are defined based on the positive and negative data pairs.
It can be seen that if there are multiple pairs corresponding to the same underlying instance (e.g., the image-text pairs that correspond to the same shape in TriCoLo), some positive pairs will appear at the non-diagonal position.
For the conventional InfoNCE loss \cite{AaronvandenOord2018RepresentationLW}, 
these pairs will be falsely treated as negative pairs since InfoNCE only selects the diagonal pairs as positive.
A simple improvement of InfoNCE for multifold data is to mask out the false negative pairs, denoted as InfoNCE+Mask.
However, InfoNCE+Mask is still single-fold-oriented and cannot simultaneously utilize information from multiple positive pairs.
Another way to improve InfoNCE is to learn soft-alignment targets similar to \cite{JunnanLi2021AlignBF,andonian2022robust} to alleviate the noisy relationships, denoted as InfoNCE+Soft.
In contrast, MFH considers both the hard and soft relationships based on the data pairs generated from the group-wise data pairing.

\begin{figure}
  \centering
  \includegraphics[width=0.9\linewidth]{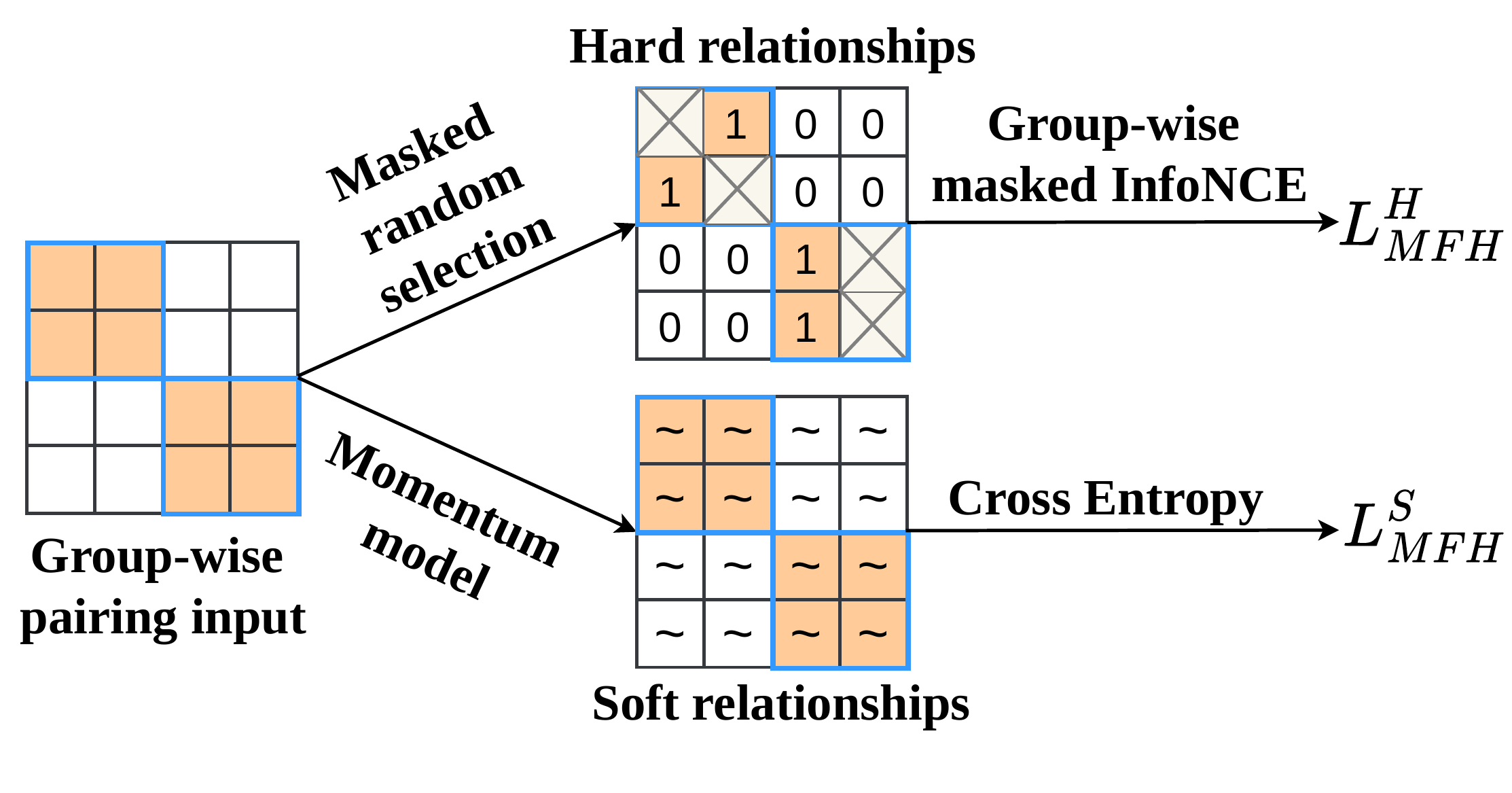}
  \vspace{-10pt}
   \caption{  
   The illustration of how MFH loss is computed.
   }
   \label{fig:mask}
   \vspace{-10pt}
\end{figure}

Specifically, the MFH loss consists of two terms: the hard relationship loss $L_{MFH}^{H}$ and the soft relationship loss $L_{MFH}^{S}$.
As shown in Figure \ref{fig:mask}, given the group-wise data pairs as input, to compute the hard relationship loss $L_{MFH}^{H}$, the first task is to decide which positive pairs should be selected for contrastive learning. 
Generally, we try to select multiple positive pairs from each group since we aim to utilize more information provided by the multifold data during one learning iteration. 
A simple selection scheme is to treat all pairs in a diagonal group as positive. 
However, such a scheme may involve some noisy (e.g., due to the mismatch between the image and text) or inconsistent (e.g., two positive pairs with similar captions but very different images) pairs, which are introduced by the Cartesian product operation.  
To alleviate the possible inconsistency of the positive pairs, we propose a \textit{masked random} pair selection scheme which randomly selects a positive pair from each row of the group matrix and masks the unselected positive pairs to avoid treating them as negative pairs.
This random selection operation can be repeated for multiple times to sample more combinations of positive pairs.
With the selected positive pairs, the hard relationship loss $L_{MFH}^{H}$ is defined as:
\begin{equation}
L_{MFH}^{H} = (L_{I\rightarrow T}^{H} + L_{T\rightarrow I}^{H}) / 2, \quad where
\end{equation}
\begin{scriptsize}
\begin{equation}
L_{I\rightarrow T}^{H} \!\!=\!\! \frac{-1}{N}\!\! \sum\limits_{x=1}\limits^{p}\! \sum\limits_{k=1}\limits^{b}\!\sum\limits_{r=1}\limits^{m}\! log \frac{exp(s(\hat{i}_{k}^{r},\hat{t}_{k}^{u}))}
{exp(s(\hat{i}_{k}^{r},\hat{t}_{k}^{u})) + \sum_{h\in \mathcal{N}_k^r}^{}\!\!exp(s(\hat{i}_{k}^{r},\hat{t}_{k}^{h}))}.
\end{equation}
\end{scriptsize}%
Here, $L_{I\rightarrow T}^{H}$ is the hard image-to-text loss defined in the form of group-wise masked InfoNCE; $p$ is the total repetition time for the masked random pair selection and $x$ is the variable to indicate the current repetition round; $N=p*b*m$ are the number of positive pairs selected; $u = random(0,n)$ is a random integer in the range $[0,n)$, representing the randomly selected column index;
$\mathcal{N}_k^r$ is the set of indices for negative samples of sample $r$ w.r.t. group $k$;
$L_{T\rightarrow I}^{H}$ is defined similarly as $L_{I\rightarrow T}^{H}$.

Although the hard relationship loss $L_{MFH}^{H}$ can guide the model to utilize more information of the multifold data, there are still some limitations:
1) all selected positive pairs are treated equally despite the similarity of some positive pairs may be low;
2) no potential positive samples can be discovered from the initial negative pairs.
To address these limitations, we incorporate soft-target learning into our MXM-CLR framework so that the soft relationships between both the positive and negative pairs can be modeled.
Specifically, we take the similar idea of ALBEF \cite{JunnanLi2021AlignBF} and use a momentum model to generate pseudo-targets for the soft relationship learning.
The momentum model is a continuously-evolving teacher that is composed of exponential-moving-average versions of the data encoders.
During training, features extracted by the base encoders of the input modalities and the momentum model are used to compute two pair-wise feature similarity matrices, while the one from the momentum model is used as the supervision to guide the training of base encoders.
Formally, the soft relationship loss $L_{MFH}^{S}$ is defined as:
\begin{equation}
L_{MFH}^{S} = (L_{I\rightarrow T}^{S} + L_{T\rightarrow I}^{S}) / 2, \quad where
\end{equation}
\begin{equation}
L_{I\rightarrow T}^{S} = CE(M_b,M_m).
\end{equation}%
Here, $M_b$ and $M_m$ are the image-text feature similarity matrix from the base model and momentum model respectively, while each element in the matrix is the temperature-controlled cosine similarity defined in Section \ref{sec:xm-clr};
$CE$ is the standard cross-entropy loss;
$L_{T\rightarrow I}^{S}$ is defined similarly as $L_{I\rightarrow T}^{S}$.

Finally, the MFH loss is defined as:
\begin{equation}
L_{MFH} = \alpha \cdot L_{MFH}^{H} + (1-\alpha) \cdot L_{MFH}^{S},
\end{equation}%
where $\alpha$ is a parameter (empirically set to 0.6) to control the weights of the hard and soft relationship losses.

\textbf{Variations of MXM-CLR.}
As a unified framework, MXM-CLR can be flexibly adapted to different forms by aggregating features of each modality and the resulting model can be used to support different downstream tasks.
For example, by aggregating multifold text features to one feature as in Figure \ref{fig:overview_loss}(c), MXM-CLR can be used to learn for more comprehensive text-to-image retrieval, e.g., use multiple sentences together to retrieval a natural image.
Similarly, aggregating multifold image features as in Figure \ref{fig:overview_loss}(d) corresponds to the text-to-shape retrieval, while the shape is represented by multi-view images.
At last, when features of both modalities are aggregated (Figure \ref{fig:overview_loss}(e)), MXM-CLR degenerates to the single-fold XM-CLR.

\section{Experiments}
\label{sec:results}

To evaluate the performance MXM-CLR, we conduct experiments on two representative datasets which are often used for cross-modal representation learning, Text2Shape \cite{chen2018text2shape} and Flickr30K \cite{PeterYoung2014FromID}.
The modalities we are interested in this paper are text, image and 3D data (3D shapes, particularly).
Meanwhile, as 3D shapes can also be represented based on the multi-view images, we only focus on learning the text and image encoders in our experiments.

\subsection{Datasets and Metrics} 
\textbf{Synthetic 3D-text dataset.} Text2Shape is a pioneering 3D-text dataset which contains 3D shapes annotated with natural language descriptions.
It consists of 6,521 chairs and 8,378 tables selected from ShapeNet \cite{chang2015shapenet}.
About five freeform natural language descriptions are collected for each shape by asking Amazon Mechanical Turkers to describe the color, shape, material and physical appearance of the object based on a pre-rendered rotating animation of the given shape.
Following TriCoLo \cite{YueRuan2022TriCoLoTC}, we employ the image-based 3D shape representation which treats a 3D shape as the aggregation of multi-view images.
We render 6 images of each textured mesh in Text2Shape using the same camera setup as in TriCoLo and use 5 text captions for each shape.
All chairs and tables data are trained together for the representation learning.

\textbf{Natural image-text dataset.} To evaluate how MXM-CLR works for natural images and their descriptions other than the synthetic images from Text2Shape, we also experiment with Flickr30K which is a widely-used image-text dataset containing 31,000 photographs of everyday activities, events and scenes collected from Flickr.
Each image is annotated with five text captions which describe the image content with different levels of specificity, e.g., from overall situation to specific actions.

\textbf{Metrics.} We adopt cross-modal retrieval as the downstream task to evaluate the performance of cross-modal representation learning.
Following prior works, the standard metrics of Recall rate (R@$k$) which considers a retrieval successful if at least one sample in the top $k$ retrievals is of the correct instance is used on Text2Shape and Flickr30K for quantitative comparisons.
In addition, we compute Normalized Discounted Cumulative Gain (NDCG) \cite{KalervoJrvelin2002CumulatedGE} which is a commonly-used information retrieval metric on Text2Shape and compare the results with related baselines.

\begin{figure*}
  \centering
  \setlength{\abovecaptionskip}{0.02cm}
  \includegraphics[width=1.0\linewidth]{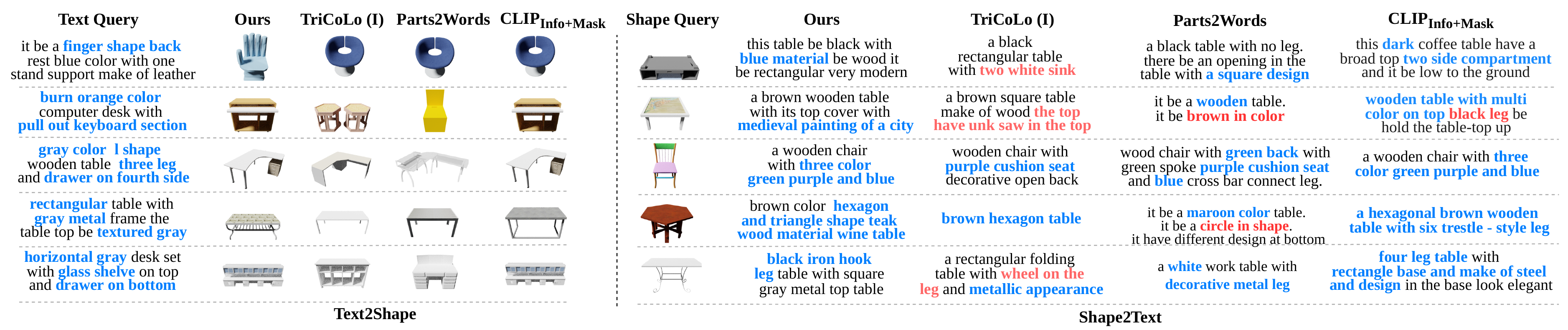}
  \caption{
  Qualitative comparison of MXM-CLR and other baselines for  $\langle$text, shape$\rangle$ retrieval tasks on Text2Shape dataset. The blue or red text are manually highlighted to indicate the parts that are matched or unmatched between the query and the retrieval result. For both tasks, MXM-CLR can obtain more fine-grained retrieval results. 
  }
  \label{fig:compare}
  \vspace{-6pt}
\end{figure*}

\subsection{Implementation Details}
\textbf{Encoders.}
We adopt the text and image encoders used in CLIP, i.e., BERT \cite{JacobDevlin2018BERTPO} for text, and ResNet101 \cite{KaimingHe2015DeepRL} or vision transformer \cite{AlexeyDosovitskiy2020AnII} for images.
For each encoder model, we use the pre-trained weights from CLIP and fine-tune the models jointly using MFH loss and other baseline losses on Text2Shape and Flickr30K.

\textbf{Training details.}
To train MXM-CLR, the encoders of text and image are initialized with the pre-trained weights from CLIP and fine-tuned for 10 epochs for each task.
The framework is implemented using PyTorch and trained on three NVIDIA RTX3090 GPUs.
The training time is depending on the type of cross-modal task, the batch size $b$ and the repetition time $p$ for masked random pair selection when computing the hard relationship loss.
For example, it takes about 0.67 hours to train MXM-CLR for the text-shape retrieval task in Table \ref{tab_text2shape_results} when repeating the selection for $p=10N_{col}$ times.
Here, $N_{col}$ is the number of columns for a group in the group matrix, e.g., the number of observations for an instance of the text modality for the $L_{I \rightarrow T}^H$ loss.
If not specifically mentioned, we also set $p=10N_{col}$ in the following experiment.
More training details and the training time for other settings are provided in the supplementary material.

\begin{table}[!t]
\centering
\setlength{\tabcolsep}{1.5pt}
\resizebox{1\columnwidth}{!}{
\begin{tabular}{ccccccccccc}
\hline
\multirow{2}{*}{Method}  &           & \multicolumn{4}{c}{Text $\Rightarrow$ Shape (Agg-Image)}                                    &           & \multicolumn{4}{c}{Shape (Agg-Image) $\Rightarrow$ Text}                                    \\ \cline{3-6} \cline{8-11} 
                        &           & R@1            & R@5            & R@10           & NDCG         &           & R@1            & R@5            & R@10           & NDCG         \\ \cline{1-1} \cline{3-6} \cline{8-11} 
Text2Shape\cite{chen2018text2shape}               &           & 0.40           & 2.37           & ---            & 1.35           &           & 0.94           & 3.69           & ---            & 0.85           \\
Y2Seq2Seq\cite{ZhizhongHan2019Y2Seq2SeqCR}                &           & 2.93           & 9.23           & ---            & 6.05           &           & 6.77           & 19.30          & ---            & 5.30           \\
TriCoLo (I)\cite{YueRuan2022TriCoLoTC}              &           & 8.28           & 24.52          & ---            & 16.52          &           & 11.91          & 32.69          & ---            & 9.37           \\
TriCoLo (V)\cite{YueRuan2022TriCoLoTC}              &           & 8.73           & 26.10          & ---            & 17.53          &           & 13.07          & 35.62          & ---            & 10.33          \\
TriCoLo (I+V)\cite{YueRuan2022TriCoLoTC}            &           & 10.25          & 29.07          & ---            & 19.85          &           & 16.33          & 42.52          & ---            & 12.73          \\ 
Parts2Words\cite{ChuanTang2021Part2WordLJ}          &           & 12.72          & 32.98          & ---            & 23.13                 &           & 19.38          & 47.17          & ---            & 15.3        \\ 
\hline

$\mathrm{CLIP_{Info}}$\cite{AlecRadford2021LearningTV} &  & 15.24& 35.72 & 47.81 & 25.66 &  & 23.12 & 49.20 & 62.71 & 17.33 \\
$\mathrm{CLIP_{Info+Soft}}$ &  & 15.15 & 36.07 & 47.65 & 25.96 &  & 24.40 & 50.94 & 62.47 & 18.12 \\
Robust-XR$^*$\cite{andonian2022robust} &  & 15.32 & 36.59 & 48.71 & 26.12  &  & 23.39 & 51.74 & 62.87 & 18.55  \\ 
$\mathrm{CLIP_{Info+Mask}}$ &  & 15.90 & 37.53 & 49.48 & 27.05 &  & 25.00 & 52.68 & 64.48 & 18.53 \\
\cline{1-1} \cline{3-6} \cline{8-11} 
\textbf{MXM-CLR} & \textbf{} & \textbf{16.83} & \textbf{39.06} & \textbf{51.44} &\textbf{ 28.38} & \textbf{} & \textbf{27.48} & \textbf{56.03} & \textbf{68.23} & \textbf{20.25} \\ \hline
\end{tabular}}
\caption{Quantitative comparison of MXM-CLR with other methods for  $\langle$text, shape$\rangle$ retrieval on Text2Shape dataset. 
ViT-B/16 is used for image encoder of the CLIP-based models and MXM-CLR, with batch size (\#images,\#texts) as (600,100) and (420,350), respectively.
NDCG is measured with top 5 results.
$^*$Results of Robust-XR are based on our own implementation as their code is not released.
}
\label{tab_text2shape_results}
\vspace{-4pt}
\end{table}

\begin{table*}[]
\centering
\resizebox{1.8\columnwidth}{!}
{
\setlength{\tabcolsep}{3pt}
\begin{tabular}{cccccccccccllccccccccccll}
\hline
\multirow{3}{*}{\begin{tabular}[c]{@{}c@{}}Loss \\ function\end{tabular}} & \multirow{3}{*}{\begin{tabular}[c]{@{}c@{}}Image \\ Encoder\end{tabular}} & \multicolumn{11}{c}{Text2Shape}                                                                                                                          &  & \multicolumn{11}{c}{Flickr30K}                                                                                                                 \\ \cline{3-13} \cline{15-25} 
                                                                          &                                                                            & \multicolumn{3}{c}{Text$\Rightarrow$Image}                   &           & \multicolumn{3}{c}{Image$\Rightarrow$Text}                   &  & \multicolumn{3}{c}{Batch}           &  & \multicolumn{3}{c}{Text$\Rightarrow$Image}                   &  & \multicolumn{3}{c}{Image$\Rightarrow$Text}                  &  & \multicolumn{3}{c}{Batch}           \\ \cline{3-5} \cline{7-9} \cline{11-13} \cline{15-17} \cline{19-21} \cline{23-25} 
                                                                          &                                                                            & R@1            & R@5            & R@10           &           & R@1            & R@5            & R@10           &  & \multicolumn{3}{c}{(imgs,texts,GB)} &  & R@1            & R@5            & R@10           &  & R@1            & R@5            & R@10          &  & \multicolumn{3}{c}{(imgs,texts,GB)} \\ \cline{1-13} \cline{15-25} 
$\mathrm{CLIP_{Info+Mask}}$                                                                   & RN101                                                                      & 15.11          & 25.87          & 33.71          &           & 22.67          & 49.29          & 61.29          &  & \multicolumn{3}{c}{(420,420,49.16)} &  & \textbf{66.86}          & 88.88          & 93.44          &  & 82.60           & 95.70           & 97.60          &  & \multicolumn{3}{c}{(400,400,44.04)} \\
MXM-CLR                                                                & RN101                                                                      & \textbf{17.75} & \textbf{29.10} & \textbf{37.52} & \textbf{} & \textbf{27.27} & \textbf{54.50} & \textbf{66.91} &  & \multicolumn{3}{c}{(420,350,47.39)} &  & 66.60           & \textbf{89.56} & \textbf{94.52} &  & \textbf{83.10}  & \textbf{96.80}  & \textbf{98.60} &  & \multicolumn{3}{c}{(80,400,20.69)}  \\ \cline{1-13} \cline{15-25}
$\mathrm{CLIP_{Info+Mask}}$                                                                      & ViT-B/16                                                                   & 16.66          & 28.36          & \textbf{36.93}          &           & 23.63          & 50.13          & 62.96          &  & \multicolumn{3}{c}{(420,420,57.24)} &  & 78.92          & 95.40           & \textbf{97.88}          &  & 92.60           & 99.00          & 99.80          &  & \multicolumn{3}{c}{(400,400,54.16)} \\
MXM-CLR                                                                 & ViT-B/16                                                                   & \textbf{17.00} & \textbf{28.50} & 36.62 &           & \textbf{24.85} & \textbf{51.31} & \textbf{63.39}  &  & \multicolumn{3}{c}{(420,350,55.64)} &  & \textbf{79.20} & \textbf{95.66} & 97.86 &  & \textbf{93.80} & \textbf{99.10} & \textbf{99.90} &  & \multicolumn{3}{c}{(80,400,23.50)}  \\ \cline{1-13} \cline{15-25} 
$\mathrm{CLIP_{Info+Mask}}$                                                                      & ViT-B/32                                                                   & 15.67          & 27.04          & 35.70           &           & 22.02          & 47.22          & 60.30           &  & \multicolumn{3}{c}{(420,420,28.01)} &  & \textbf{73.68}          & 93.00          & 96.40           &  & 87.50           & \textbf{98.00}          & 99.20          &  & \multicolumn{3}{c}{(400,400,24.07)} \\
MXM-CLR                                                                 & ViT-B/32                                                                   & \textbf{17.19} & \textbf{29.44} & \textbf{37.41}  &           & \textbf{24.51} & \textbf{51.16} & \textbf{63.67}  &  & \multicolumn{3}{c}{(420,350,26.82)} &  & 73.58  & \textbf{93.26} & \textbf{96.60}  &  & \textbf{88.00}  & \textbf{98.00} & \textbf{99.30} &  & \multicolumn{3}{c}{(80,400,17.07)}  \\ \hline
\end{tabular}}
\caption{
Evaluation of MXM-CLR trained with different image encoders for $\langle$text, image$\rangle$ retrieval task.
With comparable memory cost on Text2Shape and much less memory cost on Flickr30K, MXM-CLR generally achieves superior performance than the baseline.
Note the InfoNCE based methods construct the batch with 1:1 ratio of images and texts and MXM-CLR performs the group-wise pairing which utilizes all available observations for the same instance, i.e., the image:text ratio is 6:5 for Text2Shape and 1:5 for Flickr.
}
\label{tab_MF}
\vspace{-12pt}
\end{table*}

\subsection{Comparisons and Evaluation}

\textbf{Quantitative comparisons.} 
We first conduct quantitative experiments on Text2Shape \cite{chen2018text2shape} dataset for the text-to-shape task and compare with the existing 3D-focused cross-modal methods including Text2Shape \cite{chen2018text2shape}, Y2Seq2Seq \cite{ZhizhongHan2019Y2Seq2SeqCR}, TriCoLo \cite{YueRuan2022TriCoLoTC} and Part2Words \cite{ChuanTang2021Part2WordLJ}.
In addition, we compare with CLIP (denoted as $\mathrm{CLIP_{Info}}$) and three other strong baseline models: $\mathrm{CLIP_{Info+Soft}}$ which incorporates similar momentum-based soft learning scheme as our paper to CLIP, $\mathrm{CLIP_{Info+Mask}}$ which masks out the false negative data pairs based on the instance labels and Robust-XR \cite{andonian2022robust} which is a SOTA method for learning the soft relationships.
For Robust-XR, since their code is not released, we implement their progressive self-distillation strategy and fine-tune the pre-trained CLIP encoders for generating the results.
Following TriCoLo, we utilize the mean feature of multi-view images (denoted Agg-Image) to represent a shape and conduct cross-modal retrieval between text and shape.
From the results in Table \ref{tab_text2shape_results}, MXM-CLR outperforms the existing 3D-focused cross-modal learning methods by a significant margin and also shows notable improvements comparing to the CLIP-based strong baseline models.

\textbf{Qualitative comparisons.} 
Figure \ref{fig:compare} shows qualitative comparison between MXM-CLR with TriCoLo (I), Part2Words and $\mathrm{CLIP_{Info+Mask}}$.
The visual results show MXM-CLR can return shape/text that is more similar to the query text/shape.
More qualitative comparisons are provided in the supplementary material.

\begin{figure*}[!t]
	\centering
	\includegraphics[width=0.95\linewidth]{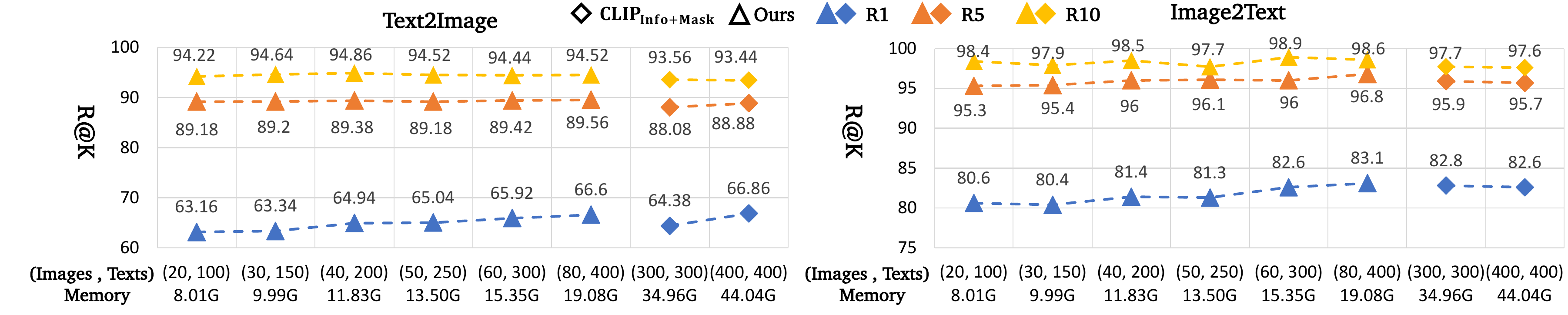}
	\caption{
	Ablation study on batch sizes of MXM-CLR for $\langle$text, image$\rangle$ retrieval task on Flickr30K.
	As a comparison, $\mathrm{CLIP_{Info+Mask}}$ results at two relatively large batch size settings are also visualized. 
	ResNet101 is used as the image encoder for this experiment.
	}
	\label{fig_batch}
	\vspace{-10pt}
\end{figure*}

\begin{table}[!t]
\centering
\resizebox{1\columnwidth}{!}{
\setlength{\tabcolsep}{1.8pt}
\begin{tabular}{cccccccccccll}
\hline
\multirow{3}{*}{\begin{tabular}[c]{@{}c@{}}Loss \\ function\end{tabular}} & \multirow{3}{*}{\begin{tabular}[c]{@{}c@{}}Image \\ Encoder\end{tabular}} & \multicolumn{11}{c}{Text2Shape}                                                                                                                                                                                                                                                  \\ \cline{3-13} 
                                                                          &                                                                            & \multicolumn{3}{c}{Text $\Rightarrow$ Agg-Image}                                                              &                      & \multicolumn{3}{c}{Agg-Image$\Rightarrow$ Text}                                                                        &  & \multicolumn{3}{c}{Batch}           \\ \cline{3-5} \cline{7-9} \cline{11-13} 
                                                                          &                                                                            & R@1                               & R@5                               & R@10                      &                      & R@1                                & R@5                                & R@10                               &  & \multicolumn{3}{c}{(imgs,texts,GB)} \\ \hline
$\mathrm{CLIP_{Info+Mask}}$                                                                    & RN101                                                                      & 15.43                             & 37.45                             & 50.00            &                      & 26.34                              & 55.43                              & 67.16                              &  & \multicolumn{3}{c}{(600,100,55.89)} \\

MXM-CLR                                                               & RN101                                                                      & \multicolumn{1}{l}{\textbf{15.84}} & \multicolumn{1}{l}{\textbf{38.41}} & \multicolumn{1}{l}{\textbf{50.54}} & \multicolumn{1}{l}{} & \multicolumn{1}{l}{\textbf{26.47}} & \multicolumn{1}{l}{\textbf{56.10}} & \multicolumn{1}{l}{\textbf{67.90}} &  & \multicolumn{3}{c}{(420,350,47.39)} \\ \hline
$\mathrm{CLIP_{Info+Mask}}$                                                                    & ViT-B/16                                                                   & 15.90                              & 37.53                             & 49.48                     &                      & 25.00                              & 52.68                              & 64.48                              &  & \multicolumn{3}{c}{(600,100,66.27)} \\
MXM-CLR                                                               & ViT-B/16                                                                   & \textbf{16.83}                       & \textbf{39.06}                    & \textbf{51.44}            &                      & \textbf{27.48}                     & \textbf{56.03}                     & \textbf{68.23}                     &  & \multicolumn{3}{c}{(420,350,55.64)} \\ \hline
$\mathrm{CLIP_{Info+Mask}}$                                                                    & ViT-B/32                                                                   & 15.47                             & 36.30                              & 48.20                      &                      & 24.60                               & 53.08                              & 64.61                              &  & \multicolumn{3}{c}{(600,100,25.25)} \\
MXM-CLR                                                               & ViT-B/32                                                                   & \textbf{16.24}                    & \textbf{37.78}                    & \textbf{49.37}            &                      & \textbf{26.41}                     & \textbf{54.76 }                   & \textbf{66.02}                     &  & \multicolumn{3}{c}{(420,350,26.82)} \\ \hline

\end{tabular}
}
\caption{
Evaluation of MXM-CLR for $\langle$text, agg-image (shape)$\rangle$ retrieval task. Since six images are aggregated into one shape feature for this task, the input ratio for image-text is 6:1 for InfoNCE-based methods.
For MXM-CLR, it computes 70 aggregated shape features from the 420 input images and pairs the 70 shapes with the 350 texts via group-wise pairing.
}

\label{tab_img_agg_mf}
\vspace{-11pt}
\end{table}

\begin{table}[!t]
\resizebox{1\columnwidth}{!}{
\setlength{\tabcolsep}{1.6pt}
\begin{tabular}{cccccccccccll}
\hline
\multirow{3}{*}{\begin{tabular}[c]{@{}c@{}}Loss \\ function\end{tabular}} & \multirow{3}{*}{\begin{tabular}[c]{@{}c@{}}Vision \\ Encoder\end{tabular}} & \multicolumn{11}{c}{Text2Shape}                                                                                                                            \\ \cline{3-13} 
                                                                          &                                                                            & \multicolumn{3}{c}{Agg-Text $\Rightarrow$ Image} &           & \multicolumn{3}{c}{Image  $\Rightarrow$ Agg-Text} &  & \multicolumn{3}{c}{Batch}            \\ \cline{3-5} \cline{7-9} \cline{11-13} 
                                                                          &                                                                            & R@1            & R@5            & R@10           &           & R@1             & R@5            & R@10           &  & \multicolumn{3}{c}{(imgs,texts,GB)}  \\ \hline
$\mathrm{CLIP_{Info+Mask}}$                                                                    & RN101                                                                      & 26.34          & 46.92          & 56.77          &           & 21.00           & 48.11          & 61.28          &  & \multicolumn{3}{c}{(256,1280,54.46)} \\

MXM-CLR                                                               & RN101                                                                      & \textbf{36.33} & \textbf{56.77} & \textbf{67.29} & \textbf{} & \textbf{24.69}   & \textbf{54.67} & \textbf{68.49} &  & \multicolumn{3}{c}{(240,200,27.59)}  \\ \hline
$\mathrm{CLIP_{Info+Mask}}$                                                                        & ViT-B/16                                                                   & 38.20           & 58.31          & 69.37          &           & 27.96           & 58.05          & 70.89          &  & \multicolumn{3}{c}{(256,1280,62.48)} \\
MXM-CLR                                                               & ViT-B/16                                                                   & \textbf{40.28} & \textbf{59.99} & \textbf{71.31} &           & \textbf{28.14}  & \textbf{59.37} & \textbf{72.54} &  & \multicolumn{3}{c}{(240,200,32.53)}  \\ \hline
$\mathrm{CLIP_{Info+Mask}}$                                                                        & ViT-B/32                                                                   & 35.46          & 55.50          & 64.88          &           & 25.25           & 53.49          & 67.27          &  & \multicolumn{3}{c}{(256,1280,42.02)} \\
MXM-CLR                                                               & ViT-B/32                                                                   & \textbf{39.95} & \textbf{59.05} & \textbf{68.36} &           & \textbf{27.49}  & \textbf{56.76} & \textbf{70.05} &  & \multicolumn{3}{c}{(240,200,15.85)}  \\ \hline
\end{tabular}
}
\caption{Evaluation of MXM-CLR for  $\langle$agg-text, image$\rangle$ retrieval task. Since five texts are aggregated into one comprehensive text feature for this task, the input ratio for image-text is 1:5 for InfoNCE-based methods.
For MXM-CLR, it computes 40 aggregated text features from the 200 input texts and pairs the 40 aggregated texts with the 240 images via group-wise pairing.
}
\label{text_agg_mf_loss}
\vspace{-11pt}
\end{table}

\textbf{Evaluation on adaptability and generalizability of MXM-CLR.}
As a unified framework, MXM-CLR can be adapted to different variations for specific tasks.
In Table \ref{tab_MF} (left), Table \ref{tab_img_agg_mf} and Table \ref{text_agg_mf_loss}, we show the quantitative results of three MXM-CLR variations on Text2Shape dataset.
The $\mathrm{CLIP_{Info+Mask}}$ which has the closest performance with MXM-CLR based on previous experiments (Table \ref{tab_text2shape_results}) is used as the baseline for comparison.
Different image encoders, including ResNet101 and different versions of vision transformers (ViT-B/16 and ViT-B/32), are tested to evaluate the generalizability of MXM-CLR.
Also, to further verify generalizability of MXM-CLR to different datasets, similar experiments are performed on Flickr30K dataset (Table \ref{tab_MF} right) for text-image retrieval.
As the batch size will affect the results for the contrastive learning methods, for a fair comparison on each task, we use the maximum batch size that a $\mathrm{CLIP_{Info+Mask}}$ model can achieve under our GPU memory limit as a reference and choose a smaller batch size for MXM-CLR.
Since the memory cost is mainly due to images, we either use the same or a smaller image number in a MXM-CLR batch than $\mathrm{CLIP_{Info+Mask}}$.

From the results, MXM-CLR can achieve the superior performance for most tasks with different image encoders, showing its adaptability and generalizability.
Moreover, while being able to learning better representations, the comparable or much smaller memory cost of MXM-CLR demonstrates the data efficiency of our method.
Note that for Flickr30K, we only want to show the generalizability of MXM-CLR and its superior performance comparing to the XM-CLR baselines, instead of aiming to beat the SOTA results based on vision-language pre-training such as \cite{wang2022image}.

\subsection{Ablation Studies}
\textbf{Ablation study on MFH loss.}
As shown in Table \ref{tab_ablation}, we investigate the contributions of different terms of MFH loss.
All experiments are conducted for the $\langle$text, image$\rangle$ retrieval task with the same batch size of 420 images and 350 texts.
To focus the evaluation on the relationship modeling schemes, we compare with another two baselines ($\mathrm{CLIP^G_{Info}}$ and $\mathrm{CLIP^G_{Info+Mask}}$) which are the group-wise adaptation of the corresponding CLIP models.
Specifically, for the group-wise paired data matrix (e.g., Fig \ref{fig:overview_loss}(b)), $\mathrm{CLIP^G_{Info}}$ sets all pairs in diagonal line as positive and other pairs in the groups as negative, while $\mathrm{CLIP^G_{Info+Mask}}$ masks out the non-diagonal pairs in the groups.
It can be seen $\mathrm{CLIP^G_{Info}}$ and $\mathrm{CLIP^G_{Info+Mask}}$ perform poorly comparing to other schemes.
This is because they only consider a fixed pairing pattern and ignore other combinations of observations which are vital for multifold cross-modal representation learning.
On the other hand, regarding all pairs in the group as positives can improve the performance to a certain degree, but it is inferior than the schemes that only select part of the pairs.
The reason is there might be some not-so-well matched pairs created by the group-wise data pairing process, e.g., a text caption describing the front side of a chair and the image is from the side view are paired based on the Cartesian product.
Hence, always training the network with all pairs may affect the contribution of other well-matched positive pairs.

Also from Table \ref{tab_ablation}, directly utilizing the hard relationship loss with the masked random positive selection scheme can achieve superior performance than the above methods since more diversified positive pair selection is helpful for MXM-CLR. 
Moreover, for most metrics, the performance of the model is positively correlated with how many times the group-wise masked random selection is repeated in an iteration, while a large repetition number will lead to a longer training time. 
On the other hand, the performance is poor if only the soft-targets generated by the momentum model are used as the supervision for MXM-CLR. 
The reason is that soft-targets alone are not enough to guide the representation learning, especially for the data with multifold observations.
When combining the soft-target learning with the hard relationship modeling, the performance can be boosted comparing to only using the hard relationship loss.

\textbf{Ablation study on batch size.}
Batch size is a critical factor for contrastive learning. Therefore, we evaluate the performance of MXM-CLR under different batch size settings for $\langle$text, image$\rangle$ retrieval task on Flickr30K.
As shown in Figure \ref{fig_batch}, the performance of the MXM-CLR increases along with the batch sizes. Besides, we also visualize the $\mathrm{CLIP_{Info+Mask}}$ result under two relatively large batch size settings.  
It can be found MXM-CLR only needs much fewer images to achieve comparable or superior performance w.r.t. $\mathrm{CLIP_{Info+Mask}}$.
This demonstrates that MXM-CLR can efficiently learn from multifold observations and mine their comprehensive relationships via the group-wise data pairing and the hybrid loss MFH.

\begin{table}[]
\resizebox{1\columnwidth}{!}{
\setlength{\tabcolsep}{1.8pt}
\begin{tabular}{cccccccccc}
\hline
\multirow{3}{*}{\begin{tabular}[c]{@{}c@{}}Relationship modeling \\ scheme\end{tabular}} & \multirow{3}{*}{\begin{tabular}[c]{@{}c@{}}Vision \\ Encoder\end{tabular}} & \multicolumn{7}{c}{Text2Shape}                                                                         & \multirow{3}{*}{\begin{tabular}[c]{@{}c@{}}Training \\ time (h)\end{tabular}} \\ \cline{3-9}
                                                                               &                                                                            & \multicolumn{3}{c}{Text$\Rightarrow$ Image}             &  & \multicolumn{3}{c}{Image $\Rightarrow$ Text}            &                                    \\ \cline{3-5} \cline{7-9}
                                                                               &                                                                            & R@1            & R@5            & R@10           &  & R@1            & R@5            & R@10           &                                    \\ \hline
$\mathrm{CLIP^G_{Info}}$                                                                       & RN101                                                                      & 8.22           & 23.04          & 32.98          &  & 9.86           & 28.71          & 41.49          & 0.62                               \\
$\mathrm{CLIP^G_{Info+Mask}}$                                                                      & RN101                                                                      & 9.84           & 26.40          & 36.94          &  & 14.19          & 35.63          & 47.91          & 0.70                               \\ 
All                                                                            & RN101                                                                      & 15.49          & 27.11          & 35.59          &  & 21.34          & 49.15          & 61.83          & 0.73                               \\ \hline
Soft                                                  & RN101                                                                      & 2.42 & 7.03          & 11.21          &  & 2.28 & 7.10 & 11.06& 0.67                               \\ 
Hard (1)                                                                 & RN101                                                                      & 15.46          & 27.13 & 35.86 &  & 24.12 & 50.51 & 62.61 & 0.80                               \\
Hard (1)+Soft                                                   & RN101                                                                      & 16.65 & 28.32          & 36.89         &  & 25.17         & 52.55          & 64.37          & 1.05                               \\

Hard ($N_{col}$)                                                              & RN101                                                                      & 16.69          & 28.71 & 37.51 &  & 25.21          & 52.70          & 65.37          & 1.01                               \\
Hard ($N_{col}$)+Soft                                                   & RN101                                                                      & 17.49 & 29.33         & 37.48         &  & 26.43          & 53.95          & 66.19          & 1.30                               \\
Hard (10$N_{col}$)                                                             & RN101                                                                      & 16.72 & 28.56          & 37.22          &  & 25.41 & 53.49 & 66.38 & 2.02                           \\ 

Hard (10$N_{col}$)+Soft                                                   & RN101                                                                      & 17.75 & 29.10          & 37.52          &  & 27.27         & 54.50         & 66.91          & 2.45                              \\

\hline
\end{tabular}}
\caption{ Ablation study on different relationship modeling schemes.
For the hard relationship modeling, the number in the bracket indicates the repetition time for the masked random pair selection when computing the $L_{I\rightarrow T}^{H}$ or $L_{T\rightarrow I}^{H}$ loss.
$\mathrm{CLIP^G_{Info}}$ and $\mathrm{CLIP^G_{Info+Mask}}$ are the group-wise adaptation of corresponding CLIP models.
}

\label{tab_ablation}
\vspace{-10pt}
\end{table}


\section{Conclusion}
In summary, we propose MXM-CLR, a unified cross-modal constrative representation learning framework which is particularly suitable for data with multifold observations.
To train MXM-CLR models, we introduce MFH, a novel multifold-aware hybrid loss, 
to simultaneously leverage multiple positive observations when computing the hard and soft relationships for the cross-modal data pairs.
Currently, we only consider the inter-modality relationships during the training. 
Inspired by works which consider relationships for both inter- and intra-modalities \cite{MohammadrezaZolfaghari2021CrossCLRCC,hong2022versatile}, incorporating the intra-modality relationships into our MXM-CLR by applying hard and soft relationship losses to data pairs constructed from the same-modal observations is a promising way to further improve the representation learning performance.
In addition, extending the MXM-CLR to other modalities such as videos (represented as multifold clips) is also an interesting future direction.

{\small
\normalem
\bibliographystyle{ieee_fullname}
\bibliography{egbib}
}

\begin{figure}[htbp]
  \centering
  \includegraphics[width=1\linewidth]{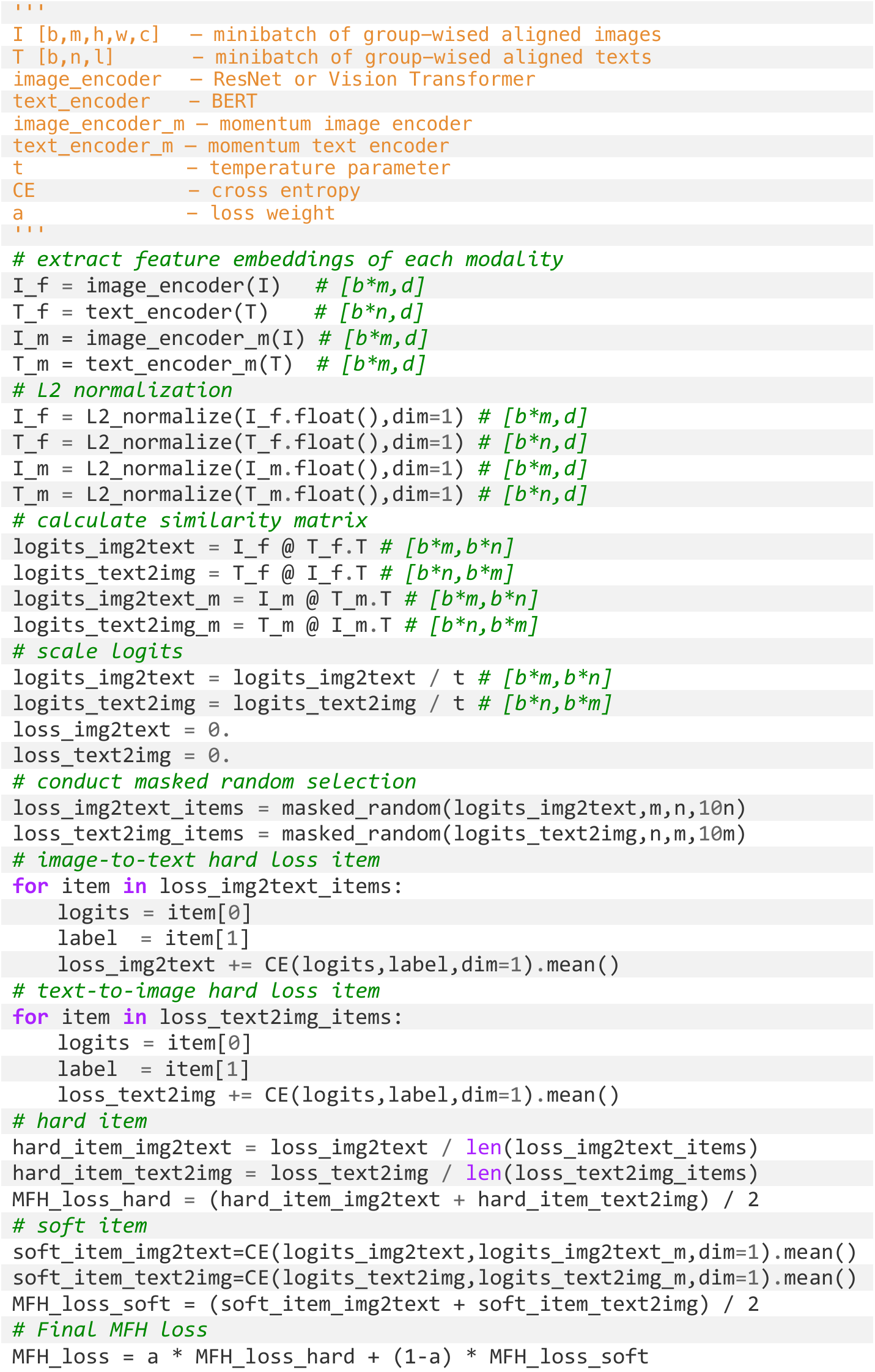}
   \caption{ Implementation of MFH loss in Python and PyTorch. 
   }
   \label{fig:mf-info}
\end{figure}

\begin{figure}[htbp]
  \centering
  \includegraphics[width=1\linewidth]{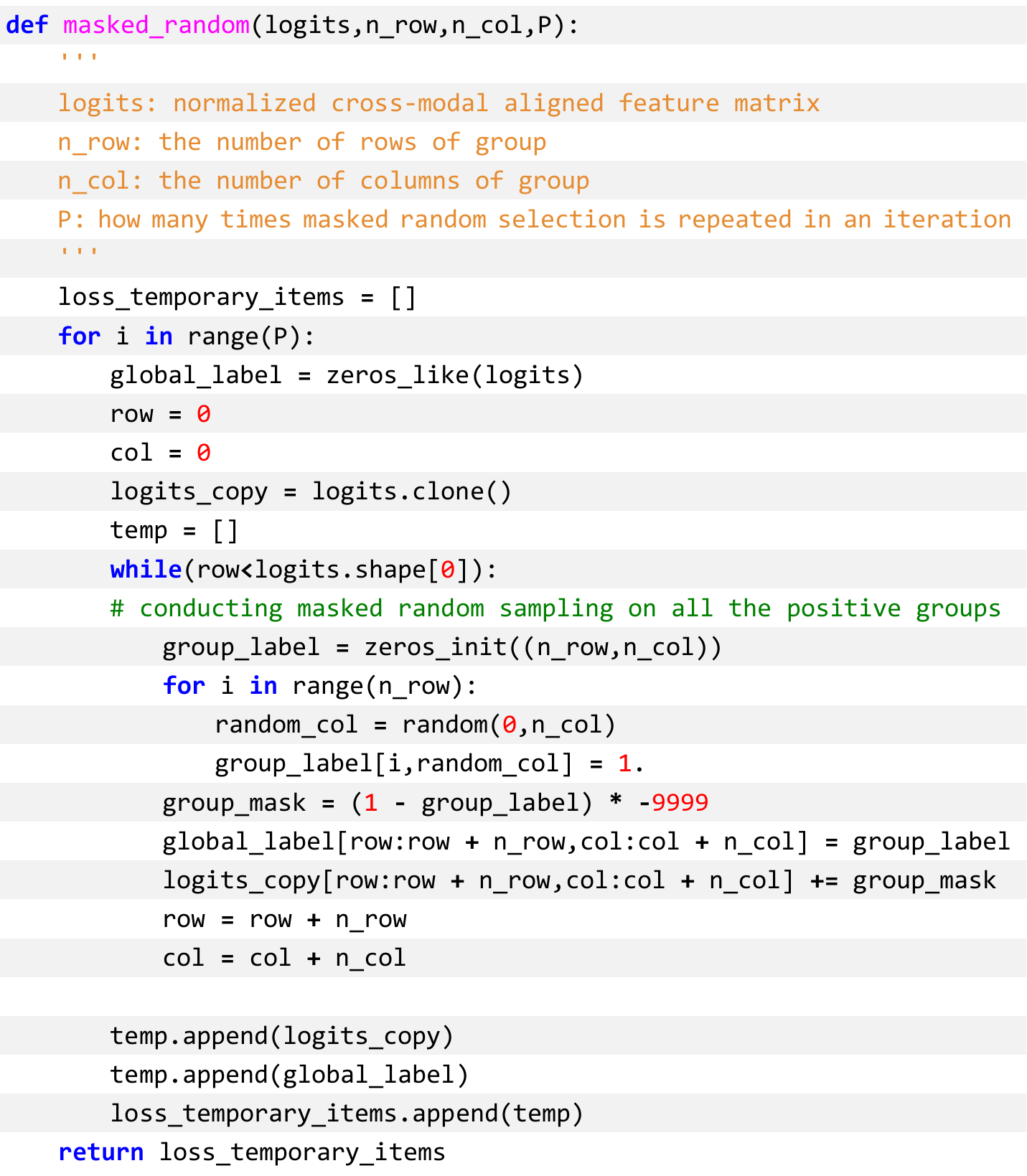}
   \caption{ Implementation of masked random positive pair selection scheme when computing the hard relationship loss. 
   }
   \label{fig:random_mf}
 
\end{figure}

\section{Appendix}
In this material, we first provide the implementation of the MFH loss and more training details.
Then, we provide more results of quantitative comparisons with related strong baselines, as well as more ablation studies on the batch sizes. 
Finally, we show more qualitative comparisons between MXM-CLR and TriCoLo \cite{YueRuan2022TriCoLoTC}, and more qualitative results of MXM-CLR on Text2Shape \cite{chen2018text2shape} and Flickr30K \cite{PeterYoung2014FromID}.

\subsection{More Technical Details}
\textbf{Implementation of MFH loss.}
We provide the code for the implementation of MFH loss in Figure \ref{fig:mf-info} and \ref{fig:random_mf}.

\begin{table}[tbp]
\resizebox{1\columnwidth}{!}{
\setlength{\tabcolsep}{2.5pt}
\begin{tabular}{cccc}
\hline
\multirow{2}{*}{Hyper-parameters} & \multicolumn{3}{c}{Image encoder}                                                                                                                                                                      \\ \cline{2-4} 
                                  & RN101                                                            & ViT-B/32                                                         & ViT-B/16                                                         \\ \hline
Number of views                   & 6                                                                & 6                                                                & 6                                                                \\
Number of captions                & 5                                                                & 5                                                                & 5                                                                \\
Batch size                        & (420, 350)                                                       & (420, 350)                                                       & (420, 350)                                                       \\
Temperature                       & 0.03                                                             & 0.03                                                             & 0.03                                                             \\
Optimizer                         & Adam                                                             & Adam                                                             & Adam                                                             \\
Adam betas                        & (0.9, 0.999)                                                     & (0.9, 0.999)                                                     & (0.9, 0.999)                                                     \\
Adam eps                          & 0.0001                                                           & 0.0001                                                           & 0.0001                                                           \\
Learning rate                     & 1e-5                                                             & 1e-6                                                             & 1e-6                                                             \\
Weight decay                      & 0.01                                                             & 0.04                                                             & 0.04                                                             \\
Scheduler                          & Cosine                                                           & Cosine                                                           & Cosine                                                           \\
Max epoches                       & 10                                                               & 10                                                               & 10                                                               \\
Random seed                       & 2022                                                             & 2022                                                             & 2022                                                             \\
\hline
\end{tabular}
}
\caption{The detailed training parameters of MXM-CLR on Text2Shape dataset for $\langle$text, agg-image (shape)$\rangle$ retrieval tasks (Table 1 and Table 3 in main paper).
}
\label{para_1}
\vspace{-4pt}
\end{table}
\begin{table}[htbp]
\resizebox{1\columnwidth}{!}{
\setlength{\tabcolsep}{2.5pt}
\begin{tabular}{cccc}
\hline
\multirow{2}{*}{Hyper-parameters} & \multicolumn{3}{c}{Image encoder}                                                                                                                                                                      \\ \cline{2-4} 
                                  & RN101                                                            & ViT-B/32                                                         & ViT-B/16                                                         \\ \hline
Number of views                   & 6                                                                & 6                                                                & 6                                                                \\
Number of captions                & 5                                                                & 5                                                                & 5                                                                \\
Batch size                        & (420, 350)                                                       & (420, 350)                                                       & (420, 350)                                                       \\
Temperature                       & 0.03                                                             & 0.03                                                             & 0.03                                                             \\
Optimizer                         & Adam                                                             & Adam                                                             & Adam                                                             \\
Adam betas                        & (0.9, 0.999)                                                     & (0.9, 0.999)                                                     & (0.9, 0.999)                                                     \\
Adam eps                          & 0.0001                                                           & 0.0001                                                           & 0.0001                                                           \\
Learning rate                     & 1e-5                                                             & 1e-6                                                             & 1e-6                                                             \\
Weight decay                      & 0.01                                                             & 0.04                                                             & 0.04                                                             \\
Scheduler                          & Cosine                                                           & Cosine                                                           & Cosine                                                           \\
Max epoches                       & 10                                                               & 10                                                               & 10                                                               \\
Random seed                       & 2022                                                             & 2022                                                             & 2022                                                             \\
\hline
\end{tabular}
}
\caption{The detailed training parameters of MXM-CLR on Text2Shape dataset for $\langle$text, image$\rangle$ retrieval tasks (Table 2 Left in main paper).
}
\vspace{-4pt}
\label{para_2}
\end{table}

\begin{table}[htbp]
\resizebox{1\columnwidth}{!}{
\setlength{\tabcolsep}{2.5pt}
\begin{tabular}{cccc}
\hline
\multirow{2}{*}{Hyper-parameters} & \multicolumn{3}{c}{Image encoder}                                                                                                                                                                      \\ \cline{2-4} 
                                  & RN101                                                            & ViT-B/32                                                         & ViT-B/16                                                         \\ \hline
Number of views                   & 6                                                                & 6                                                                & 6                                                                \\
Number of captions                & 5                                                                & 5                                                                & 5                                                                \\
Batch size                        & (240, 200)                                                       & (240, 200)                                                       & (240, 200)                                                       \\
Temperature                       & 0.03                                                             & 0.03                                                             & 0.03                                                             \\
Optimizer                         & Adam                                                             & Adam                                                             & Adam                                                             \\
Adam betas                        & (0.9, 0.999)                                                     & (0.9, 0.999)                                                     & (0.9, 0.999)                                                     \\
Adam eps                          & 0.0001                                                           & 0.0001                                                           & 0.0001                                                           \\
Learning rate                     & 1e-5                                                             & 1e-6                                                             & 1e-6                                                             \\
Weight decay                      & 0.01                                                             & 0.04                                                             & 0.04                                                             \\
Scheduler                          & Cosine                                                           & Cosine                                                           & Cosine                                                           \\
Max epoches                       & 10                                                               & 10                                                               & 10                                                               \\
Random seed                       & 2022                                                             & 2022                                                             & 2022                                                             \\
\hline
\end{tabular}
}
\caption{The detailed training parameters of MXM-CLR on Text2Shape dataset for $\langle$agg-text, image$\rangle$ retrieval tasks (Table 4 in main paper).
}
\label{para_3}
\end{table}
\begin{table}[htbp]
\resizebox{1\columnwidth}{!}{
\setlength{\tabcolsep}{2.5pt}
\begin{tabular}{cccc}
\hline
\multirow{2}{*}{Hyper-parameters} & \multicolumn{3}{c}{Image encoder}                                                                                                                                                                      \\ \cline{2-4} 
                                  & RN101                                                            & ViT-B/32                                                         & ViT-B/16                                                         \\ \hline
Number of views                   & 1                                                                & 1                                                                & 1                                                                \\
Number of captions                & 5                                                                & 5                                                                & 5                                                                \\
Batch size                        & (80, 400)                                                        & (80, 400)                                                        & (80, 400)                                                      \\
Temperature                       & 0.03                                                             & 0.03                                                             & 0.03                                                             \\
Optimizer                         & Adam                                                             & Adam                                                             & Adam                                                             \\
Adam betas                        & (0.9, 0.999)                                                     & (0.9, 0.999)                                                     & (0.9, 0.999)                                                     \\
Adam eps                          & 0.0001                                                           & 0.0001                                                           & 0.0001                                                           \\
Learning rate                     & 1e-5                                                             & 1e-6                                                             & 1e-6                                                             \\
Weight decay                      & 0.01                                                             & 0.01                                                             & 0.01                                                             \\
Scheduler                          & Cosine                                                           & Cosine                                                           & Cosine                                                           \\
Max epoches                       & 10                                                               & 10                                                               & 10                                                               \\
Random seed                       & 2022                                                             & 2022                                                             & 2022                                                             \\
\hline
\end{tabular}
}
\caption{The detailed training parameters of MXM-CLR on Flickr30K dataset for $\langle$text, image$\rangle$ retrieval tasks  (Table 2 Right in main paper).
}
\label{para_4}
\end{table}

\textbf{More training details.}
The images from Text2Shape are rendered at $224\times224$ resolution.
All images of Text2Shape and Flickr30K are resized to $256\times256$ and transformed by conventional data augmentation techniques including random crop, random horizontal flip, random vertical flip and random rotation.
We use the Adam optimizer with momentum 0.9.
The initial learning rate is 1e-5 and updated by the cosine annealing scheduler.
In the main paper, we provide the training time of MXM-CLR for $\langle$text, image$\rangle$ and $\langle$text, agg-image (shape)$\rangle$ retrieval tasks on Text2Shape dataset.
For the $\langle$agg-text, image$\rangle$ retrieval task, it takes approximately 0.71 hours to train the MXM-CLR with ViT-B/16 \cite{AlexeyDosovitskiy2020AnII} image encoder and BERT text encoder \cite{JacobDevlin2018BERTPO}.
It can be found based on the pre-trained weights of the image and text encoders, the training of MXM-CLR, i.e., fine-tuning the pre-trained weights, is quite efficient.
The hyper-parameters for training MXM-CLR models for different tasks on Text2Shape and Flickr30K are provided in Table \ref{para_1}, \ref{para_2}, \ref{para_3} and \ref{para_4}.
For these experiments, we also set repetition time for masked random pair selection when computing the hard relationship loss as $p=10N_{Col}$.

\begin{table*}[]
\centering
\resizebox{2\columnwidth}{!}
{
\setlength{\tabcolsep}{3pt}
\begin{tabular}{cccccccccccllccccccccccll}
\hline
\multirow{3}{*}{\begin{tabular}[c]{@{}c@{}}Loss \\ function\end{tabular}} & \multirow{3}{*}{\begin{tabular}[c]{@{}c@{}}Image \\ Encoder\end{tabular}} & \multicolumn{11}{c}{Text2Shape}                                                                                                                          &  & \multicolumn{11}{c}{Flickr30K}                                                                                                                 \\ \cline{3-13} \cline{15-25} 
                                                                          &                                                                            & \multicolumn{3}{c}{Text$\Rightarrow$Image}                   &           & \multicolumn{3}{c}{Image$\Rightarrow$Text}                   &  & \multicolumn{3}{c}{Batch}           &  & \multicolumn{3}{c}{Text$\Rightarrow$Image}                   &  & \multicolumn{3}{c}{Image$\Rightarrow$Text}                  &  & \multicolumn{3}{c}{Batch}           \\ \cline{3-5} \cline{7-9} \cline{11-13} \cline{15-17} \cline{19-21} \cline{23-25} 
                                                                          &                                                                            & R@1            & R@5            & R@10           &           & R@1            & R@5            & R@10           &  & \multicolumn{3}{c}{(imgs,texts,GB)} &  & R@1            & R@5            & R@10           &  & R@1            & R@5            & R@10          &  & \multicolumn{3}{c}{(imgs,texts,GB)} \\ \cline{1-13} \cline{15-25} 
$\mathrm{CLIP_{Info+Soft}}$                                                                   & RN101                                                                      & 12.36          & 22.62          & 30.38          &           & 18.07          & 41.57          & 53.27          &  & \multicolumn{3}{c}{(420,420,49.16)} &  & 63.40          & 87.82          & 93.84          &  & 78.50           & 94.40           & 97.60          &  & \multicolumn{3}{c}{(400,400,44.04)} \\
Robust-XR \cite{andonian2022robust}                                                                  & RN101                                                                      & 12.56          & 23.41          & 31.12          &           & 18.07          & 41.57          & 53.27          &  & \multicolumn{3}{c}{(420,420,49.16)} &  & 64.04          & 87.84          & 93.00         &  & 80.30           & 94.30           & 97.20          &  & \multicolumn{3}{c}{(400,400,44.04)} \\
MXM-CLR                                                                & RN101                                                                      & \textbf{17.75} & \textbf{29.10} & \textbf{37.52} & \textbf{} & \textbf{27.27} & \textbf{54.50} & \textbf{66.91} &  & \multicolumn{3}{c}{(420,350,47.39)} &  & \textbf{66.60}          & \textbf{89.56} & \textbf{94.52} &  & \textbf{83.10}  & \textbf{96.80}  & \textbf{98.60} &  & \multicolumn{3}{c}{(80,400,20.69)}  \\ \cline{1-13} \cline{15-25}
$\mathrm{CLIP_{Info+Soft}}$                                                                      & ViT-B/16                                                                   & 16.04          & 27.95          & 36.38          &           & 23.45          & 48.25          & 60.20          &  & \multicolumn{3}{c}{(420,420,57.24)} &  & \textbf{79.80}          & \textbf{95.70}           & 97.80          &  & 92.90           & 98.90          & 99.60          &  & \multicolumn{3}{c}{(400,400,54.16)} \\
Robust-XR \cite{andonian2022robust}                                                                     & ViT-B/16                                                                   & 15.75          & 28.03          & 36.51          &           & 22.91          & 49.19          & 62.04          &  & \multicolumn{3}{c}{(420,420,57.24)} &  & 78.82          & 95.14           & 97.66          &  & 92.90           & 98.90          & 99.70          &  & \multicolumn{3}{c}{(400,400,54.16)} \\
MXM-CLR                                                                 & ViT-B/16                                                                   & \textbf{17.00} & \textbf{28.50} & \textbf{36.62} &           & \textbf{24.85} & \textbf{51.31} & \textbf{63.39}  &  & \multicolumn{3}{c}{(420,350,55.64)} &  & 79.20 & 95.66 & \textbf{97.86} &  & \textbf{93.80} & \textbf{99.10} & \textbf{99.90} &  & \multicolumn{3}{c}{(80,400,23.50)}  \\ \cline{1-13} \cline{15-25} 
$\mathrm{CLIP_{Info+Soft}}$                                                                      & ViT-B/32                                                                   & 15.12          & 27.27          & 35.00           &           & 21.30          & 46.24          & 58.41           &  & \multicolumn{3}{c}{(420,420,28.01)} &  & \textbf{73.62}          & 92.96          & 96.38           &  & 87.10           & 97.90          & 99.20          &  & \multicolumn{3}{c}{(400,400,24.07)} \\
Robust-XR \cite{andonian2022robust}                                                                       & ViT-B/32                                                                   & 15.06          & 26.89          & 35.12           &           & 21.67          & 46.98          & 59.14           &  & \multicolumn{3}{c}{(420,420,28.01)} &  & 73.18          & 92.54          & 96.22           &  & 86.90           & 97.40          & \textbf{99.40}          &  & \multicolumn{3}{c}{(400,400,24.07)} \\

MXM-CLR                                                                 & ViT-B/32                                                                   & \textbf{17.19} & \textbf{29.44} & \textbf{37.41}  &           & \textbf{24.51} & \textbf{51.16} & \textbf{63.67}  &  & \multicolumn{3}{c}{(420,350,26.82)} &  & 73.58  & \textbf{93.26} & \textbf{96.60}  &  & \textbf{88.00}  & \textbf{98.00} & 99.30 &  & \multicolumn{3}{c}{(80,400,17.07)}  \\ \hline
\end{tabular}}
\caption{
\wy{
Evaluation of MXM-CLR trained with different image encoders for $\langle$text, image$\rangle$ retrieval task.
With comparable memory cost on Text2Shape and much less memory cost on Flickr30K, MXM-CLR generally achieves superior performance than the baselines.
Note the $\mathrm{CLIP_{Info+Soft}}$ and Robust-XR \cite{andonian2022robust}  construct the batch with 1:1 ratio of images and texts and MXM-CLR performs the group-wise pairing which utilizes all available observations for the same instance, i.e., the image:text ratio is 6:5 for Text2Shape and 1:5 for Flickr.
}
}

\label{tab_6}
\vspace{-12pt}
\end{table*}

\subsection{More Quantitative Results}

\wy{\textbf{More comparison with strong baselines.}
To further demonstrate the performance of our method, we perform more experiments to compare two strong CLIP baseline models which consider soft relationships for the data pairs. As shown in Table \ref{tab_6}, \ref{tab_7} and \ref{tab_8}, MXM-CLR outperforms $\mathrm{CLIP_{Info+Soft}}$ and Robust-XR \cite{andonian2022robust} in most tasks, demonstrating its superior performance on more comprehensive representation learning.
}


\begin{table}[!t]
\centering
\resizebox{1\columnwidth}{!}{
\setlength{\tabcolsep}{1.8pt}
\begin{tabular}{cccccccccccll}
\hline
\multirow{3}{*}{\begin{tabular}[c]{@{}c@{}}Loss \\ function\end{tabular}} & \multirow{3}{*}{\begin{tabular}[c]{@{}c@{}}Image \\ Encoder\end{tabular}} & \multicolumn{11}{c}{Text2Shape}                                                                                                                                                                                                                                                  \\ \cline{3-13} 
                                                                          &                                                                            & \multicolumn{3}{c}{Text $\Rightarrow$ Agg-Image}                                                              &                      & \multicolumn{3}{c}{Agg-Image$\Rightarrow$ Text}                                                                        &  & \multicolumn{3}{c}{Batch}           \\ \cline{3-5} \cline{7-9} \cline{11-13} 
                                                                          &                                                                            & R@1                               & R@5                               & R@10                      &                      & R@1                                & R@5                                & R@10                               &  & \multicolumn{3}{c}{(imgs,texts,GB)} \\ \hline
$\mathrm{CLIP_{Info+Soft}}$                                                                    & RN101                                                                      & 10.03                             & 27.34                             & 38.15            &                      & 12.86                              & 35.38                              & 47.18                              &  & \multicolumn{3}{c}{(600,100,55.89)} \\
Robust-XR \cite{andonian2022robust}                                                                     & RN101                                                                      & 15.35                             & 37.79                             & 49.95            &                      & \textbf{27.07}                              & 54.35                              & \textbf{67.90 }                             &  & \multicolumn{3}{c}{(600,100,55.89)} \\

MXM-CLR                                                               & RN101                                                                      & \multicolumn{1}{l}{\textbf{15.84}} & \multicolumn{1}{l}{\textbf{38.41}} & \multicolumn{1}{l}{\textbf{50.54}} & \multicolumn{1}{l}{} & \multicolumn{1}{l}{26.47} & \multicolumn{1}{l}{\textbf{56.10}} & \multicolumn{1}{l}{\textbf{67.90}} &  & \multicolumn{3}{c}{(420,350,47.39)} \\ \hline
$\mathrm{CLIP_{Info+Soft}}$                                                                    & ViT-B/16                                                                   & 15.22                              & 36.02                             & 47.38                     &                      & 24.46                              & 50.67                              & 62.20                              &  & \multicolumn{3}{c}{(600,100,66.27)} \\

Robust-XR \cite{andonian2022robust}                                                                     & ViT-B/16                                                                   & 14.08                              & 35.52                             & 47.20                     &                       & 25.07                              & 53.28                             & 65.15                              &  & \multicolumn{3}{c}{(600,100,66.27)} \\

MXM-CLR                                                               & ViT-B/16                                                                   & \textbf{16.83}                       & \textbf{39.06}                    & \textbf{51.44}            &                      & \textbf{27.48}                     & \textbf{56.03}                     & \textbf{68.23}                     &  & \multicolumn{3}{c}{(420,350,55.64)} \\ \hline
$\mathrm{CLIP_{Info+Soft}}$                                                                    & ViT-B/32                                                                   & 13.94                             & 34.00                              & 46.08                      &                      & 24.19                               & 49.53                              & 61.80                              &  & \multicolumn{3}{c}{(600,100,25.25)} \\
Robust-XR \cite{andonian2022robust}                                                                     & ViT-B/32                                                                   &  14.12                            & 34.56                              &    46.87                    &                      &  22.38                             & 48.79                              & 60.25                              &  & \multicolumn{3}{c}{(600,100,25.25)} \\
MXM-CLR                                                               & ViT-B/32                                                                   & \textbf{16.24}                    & \textbf{37.78}                    & \textbf{49.37}            &                      & \textbf{26.41}                     & \textbf{54.76 }                   & \textbf{66.02}                     &  & \multicolumn{3}{c}{(420,350,26.82)} \\ \hline

\end{tabular}
}
\caption{
\wy{Evaluation of MXM-CLR for $\langle$text, agg-image (shape)$\rangle$ retrieval task. Since six images are aggregated into one shape feature for this task, the input ratio for image-text is 6:1 for  $\mathrm{CLIP_{Info+Soft}}$ and Robust-XR \cite{andonian2022robust} .
For MXM-CLR, it computes 70 aggregated shape features from the 420 input images and pairs the 70 shapes with the 350 texts via group-wise pairing.}
}

\label{tab_7}
\vspace{-11pt}
\end{table}

\begin{table}[!t]
\resizebox{1\columnwidth}{!}{
\setlength{\tabcolsep}{1.6pt}
\begin{tabular}{cccccccccccll}
\hline
\multirow{3}{*}{\begin{tabular}[c]{@{}c@{}}Loss \\ function\end{tabular}} & \multirow{3}{*}{\begin{tabular}[c]{@{}c@{}}Vision \\ Encoder\end{tabular}} & \multicolumn{11}{c}{Text2Shape}                                                                                                                            \\ \cline{3-13} 
                                                                          &                                                                            & \multicolumn{3}{c}{Agg-Text $\Rightarrow$ Image} &           & \multicolumn{3}{c}{Image  $\Rightarrow$ Agg-Text} &  & \multicolumn{3}{c}{Batch}            \\ \cline{3-5} \cline{7-9} \cline{11-13} 
                                                                          &                                                                            & R@1            & R@5            & R@10           &           & R@1             & R@5            & R@10           &  & \multicolumn{3}{c}{(imgs,texts,GB)}  \\ \hline
$\mathrm{CLIP_{Info+Soft}}$                                                                    & RN101                                                                      & 23.59          & 41.35          & 52.74          &           & 18.57           & 44.31          & 58.02          &  & \multicolumn{3}{c}{(256,1280,54.46)} \\
Robust-XR \cite{andonian2022robust}                                                                     & RN101                                                                      & 29.56          & 48.73          & 59.05          &           & 19.83           & 48.79          & 62.48          &  & \multicolumn{3}{c}{(256,1280,54.46)} \\
MXM-CLR                                                               & RN101                                                                      & \textbf{36.33} & \textbf{56.77} & \textbf{67.29} & \textbf{} & \textbf{24.69}   & \textbf{54.67} & \textbf{68.49} &  & \multicolumn{3}{c}{(240,200,27.59)}  \\ \hline
$\mathrm{CLIP_{Info+Soft}}$                                                                        & ViT-B/16                                                                   & 37.66           & 58.44          & 69.77          &           & \textbf{28.20}           & \textbf{59.53}         & 72.26          &  & \multicolumn{3}{c}{(256,1280,62.48)} \\
Robust-XR \cite{andonian2022robust}                                                                & ViT-B/16                                                                   & 39.14 & 58.71 & 69.03 &           & 28.05  & 57.94 & 70.63 &  & \multicolumn{3}{c}{(240,200,32.53)}  \\ 
MXM-CLR                                                               & ViT-B/16                                                                   & \textbf{40.28} & \textbf{59.99} & \textbf{71.31} &           & 28.14  & 59.37 & \textbf{72.54} &  & \multicolumn{3}{c}{(240,200,32.53)}  \\ \hline
$\mathrm{CLIP_{Info+Soft}}$                                                                        & ViT-B/32                                                                   & 36.93          & 57.03          & 67.29          &           & 25.63           & 54.41          & 67.88          &  & \multicolumn{3}{c}{(256,1280,42.02)} \\
Robust-XR \cite{andonian2022robust}                                                                & ViT-B/32                                                                   & 35.45 & 55.50 & 65.82 &           & 26.20  & 53.88 & 67.32 &  & \multicolumn{3}{c}{(240,200,15.85)}  \\ 
MXM-CLR                                                               & ViT-B/32                                                                   & \textbf{39.95} & \textbf{59.05} & \textbf{68.36} &           & \textbf{27.49}  & \textbf{56.76} & \textbf{70.05} &  & \multicolumn{3}{c}{(240,200,15.85)}  \\ \hline
\end{tabular}
}
\caption{
\wy{
Evaluation of MXM-CLR for  $\langle$agg-text, image$\rangle$ retrieval task. Since five texts are aggregated into one comprehensive text feature for this task, the input ratio for image-text is 1:5 for  $\mathrm{CLIP_{Info+Soft}}$ and Robust-XR \cite{andonian2022robust} .
For MXM-CLR, it computes 40 aggregated text features from the 200 input texts and pairs the 40 aggregated texts with the 240 images via group-wise pairing.
}
}
\label{tab_8}
\vspace{-11pt}
\end{table}

\begin{table}[]
\resizebox{1\columnwidth}{!}{
\setlength{\tabcolsep}{1.8pt}
\begin{tabular}{ccccccccccc}
\hline
\multicolumn{3}{c}{Batch}                                            &  & \multicolumn{7}{c}{Flickr30K}                                                                         \\ \hline
\multirow{2}{*}{image} & \multirow{2}{*}{text} & \multirow{2}{*}{GB} &  & \multicolumn{3}{c}{Text $\Rightarrow$ Image}     &  & \multicolumn{3}{c}{Image $\Rightarrow$ Text}     \\ \cline{5-7} \cline{9-11} 
                       &                       &                     &  & R@1            & R@5            & R@10           &  & R@1            & R@5            & R@10           \\ \cline{1-3} \cline{5-11} 
20                     & 100                   & 8.01                &  & 63.16          & 89.18          & 94.22          &  & 80.60          & 95.30          & 98.40          \\
30                     & 150                   & 9.99                &  & 63.34          & 89.20          & 94.64          &  & 80.40          & 95.40          & 97.90          \\
40                     & 200                   & 11.83               &  & 64.94          & 89.38          & 94.86          &  & 81.40          & 96.00          & 98.50          \\
50                     & 250                   & 13.50               &  & 65.04          & 89.18          & 94.52          &  & 81.30          & 96.10          & 97.70          \\
60                     & 300                   & 15.35               &  & 65.92          & 89.42          & 94.44          &  & 82.60          & 96.00          & 98.90          \\
80                     & 400                   & 19.08               &  & 66.60          & 89.56          & 94.52          &  & 83.10          & 96.80          & 98.60          \\ \hline
100                    & 500                   & 22.53               &  & 67.00          & 89.94          & 94.98          &  & 82.80          & 96.50          & 98.70          \\
150                    & 750                   & 32.85               &  & 67.16          & 90.30          & \textbf{95.02} &  & 83.30          & 96.40          & 98.50          \\
200                    & 1000                  & 42.34               &  & \textbf{68.98} & \textbf{90.56} & 94.70          &  & \textbf{86.40} & \textbf{97.40} & \textbf{99.00} \\ \hline
\end{tabular}
}
\caption{ Evaluation on larger batch sizes of MXM-CLR for $\langle$ text, image $\rangle$ retrieval tasks on Flickr30K. ResNet101 is adopted as the image encoder for this experiment.
The results above the horizontal line have been reported in Figure 7 of the main paper.
}
\label{tab_large_batch}
\end{table}
\textbf{More evaluations on batch sizes in MXM-CLR.}
We perform additional experiments on Flickr30K by training MXM-CLR with larger batch size settings. 
From Table \ref{tab_large_batch}, as expected, MXM-CLR can generally achieve increasing performance as the batch size increases.
Note that we only compare the results for batch size up to (200,1000) image-text pairs due to the limit of our GPU resources.
Whether the performance could be further improved by larger batch sizes is worth investigating in the future.

\subsection{More Qualitative Results}

\begin{figure*}[htbp]
  \centering
  \includegraphics[width=0.95\linewidth]{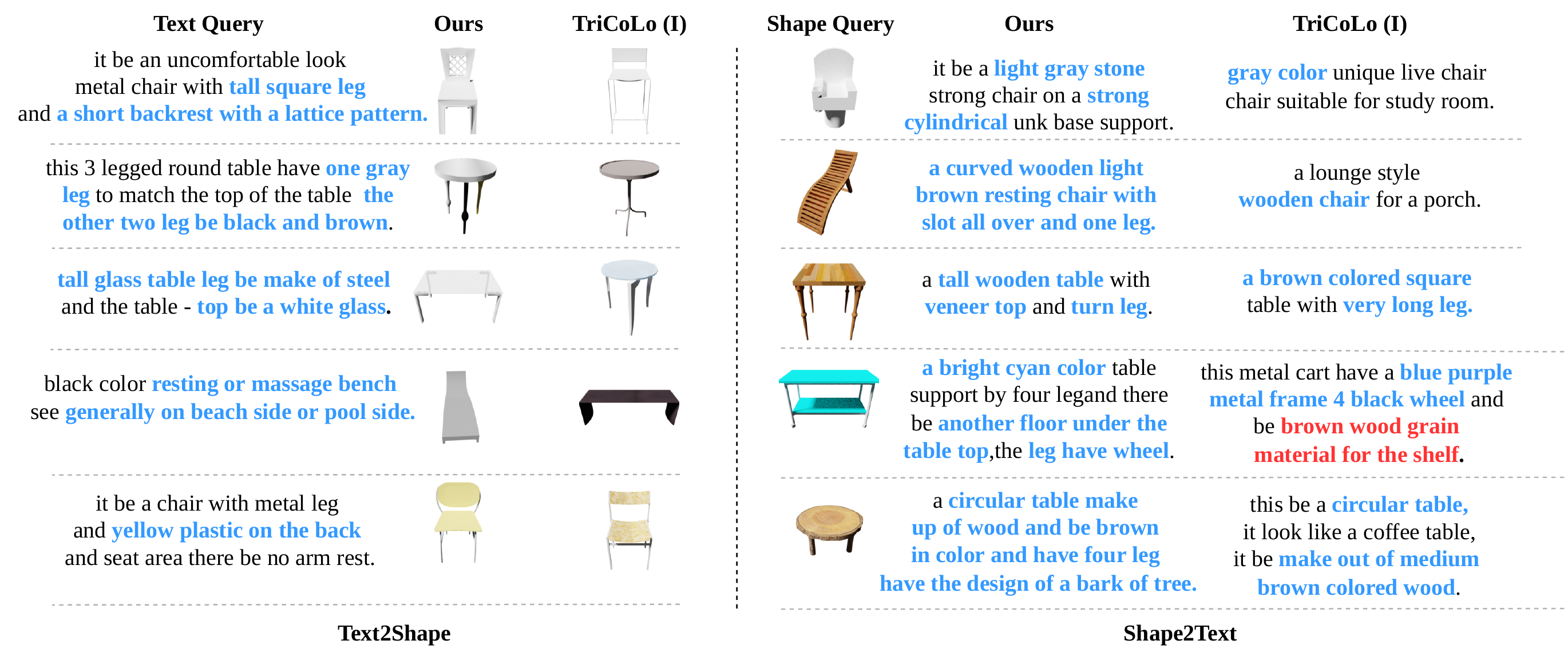}
   \caption{ More comparisons between MXM-CLR and TriCoLo (I)\cite{YueRuan2022TriCoLoTC} for cross-modal retrieval on Text2Shape\cite{chen2018text2shape} dataset. The blue
or red texts are manually highlighted to indicate the parts that are matched or unmatched between the query and the retrieval result.
   }
   \label{fig:fig_compare}
\end{figure*}

\begin{figure*}[htbp]
  \centering
  \includegraphics[width=0.93\linewidth]{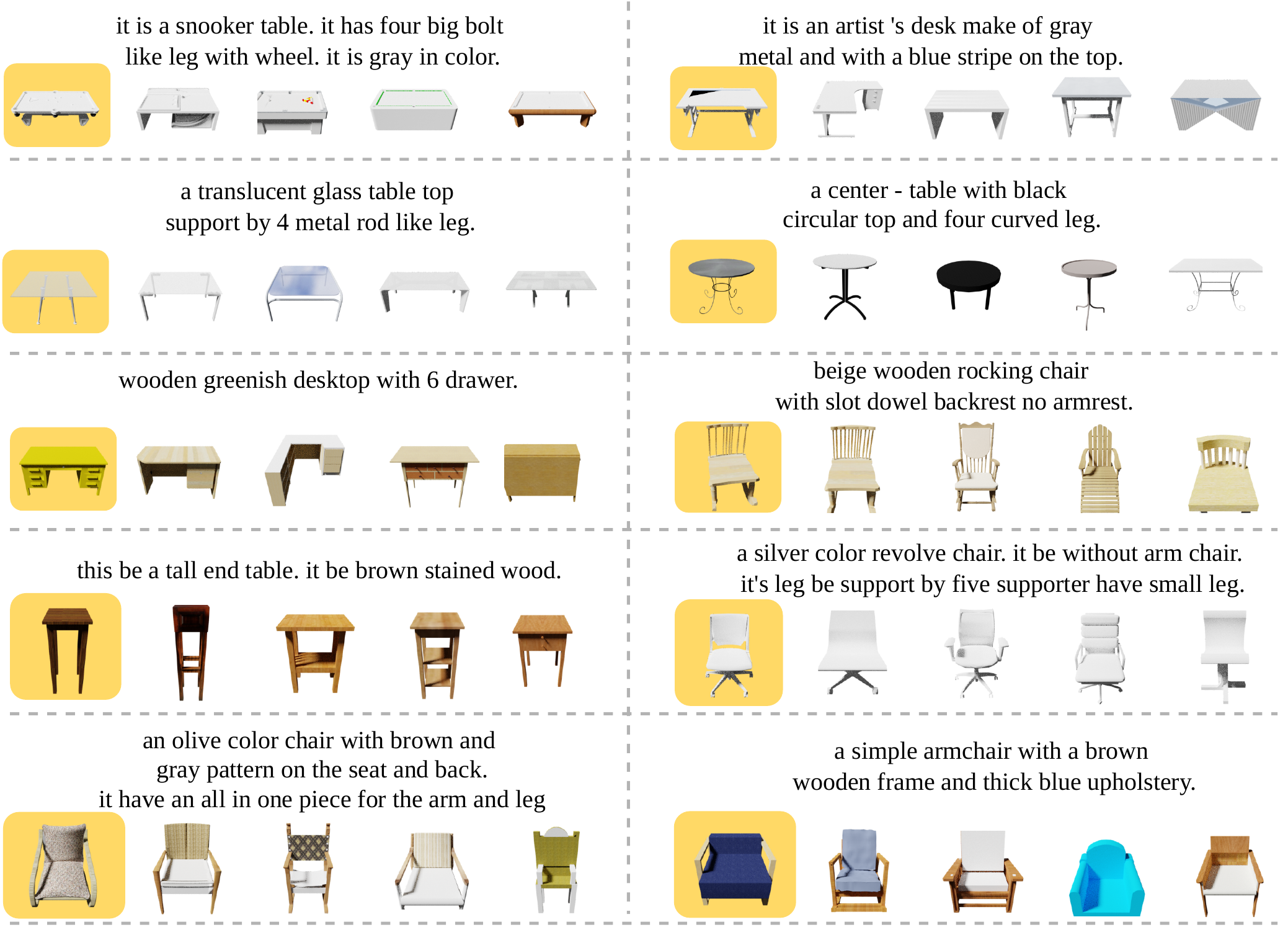}
   \caption{Examples of text-to-shape retrieval results (top 5) of MXM-CLR on Text2Shape\cite{chen2018text2shape} dataset. The shape with yellow background  indicates the ground truth shape. In most cases, besides the top 1 result, other results also consist similar content to the query text. 
   }
   \label{fig:t2i}
\end{figure*}

\begin{figure*}[htbp]
  \centering
  \includegraphics[width=0.85\linewidth]{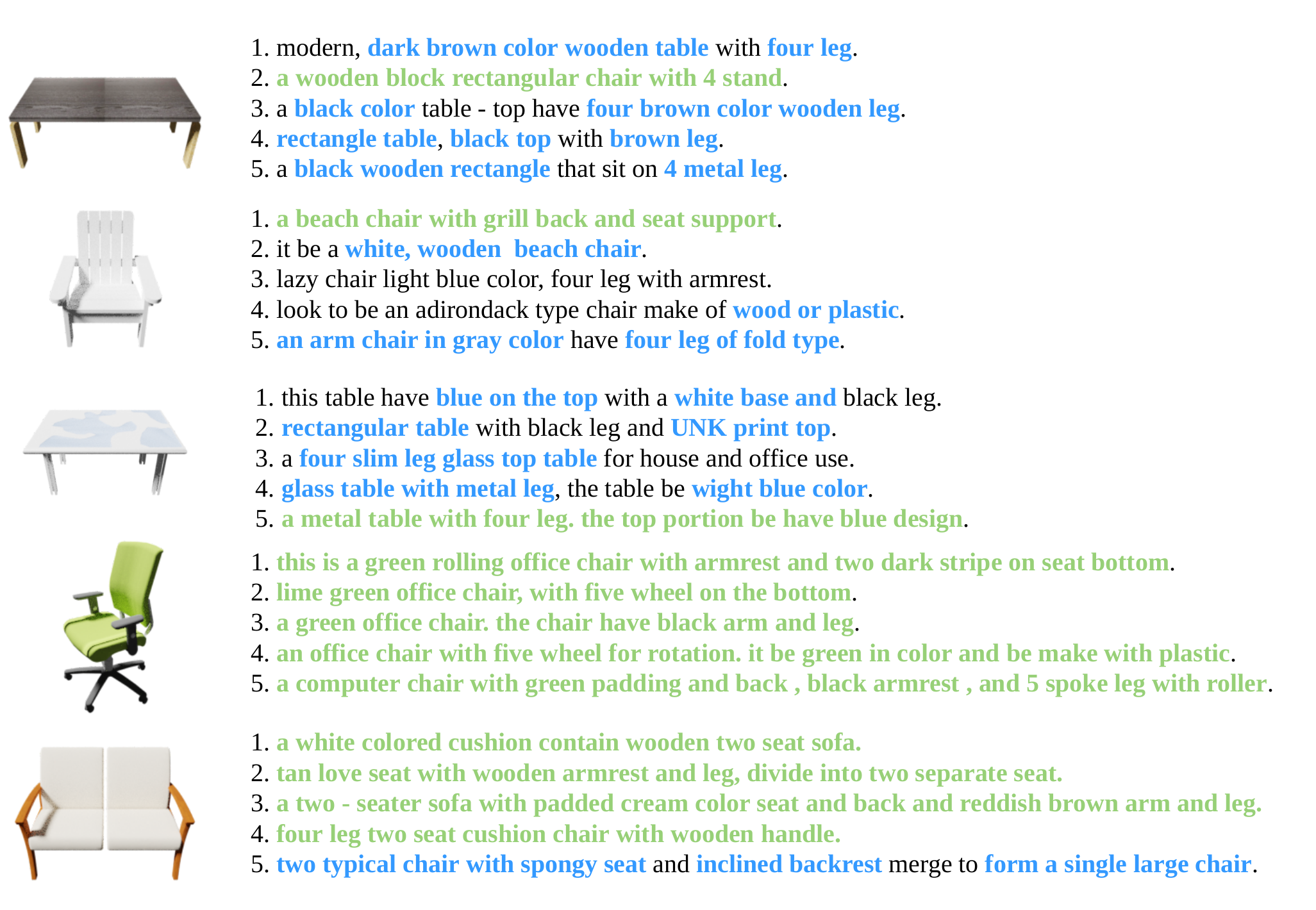}
  \vspace{-2pt}
   \caption{ Examples of shape-to-text retrieval results (top 5) of MXM-CLR on Text2Shape\cite{chen2018text2shape} dataset. The green texts are the ground truth captions and the blue texts are manually annotated to indicate the matched content w.r.t the query.
   }
   \label{fig:i2t}
\end{figure*}

\textbf{More results on Text2Shape dataset.} 
In Figure \ref{fig:fig_compare}, we show more comparisons on retrieval results of MXM-CLR and TriCoLo (I)\cite{YueRuan2022TriCoLoTC} on Text2Shape dataset \cite{chen2018text2shape}. 
MXM-CLR is able to achieve more fine-grained retrieval compared to TriCoLo, due to its better learned cross-model representations. 
Furthermore, we provide more results of MXM-CLR on $\langle$text, shape$\rangle$ retrieval tasks in Figure \ref{fig:t2i} and \ref{fig:i2t}.

\begin{figure*}[htbp]
  \centering
  \includegraphics[width=0.86\linewidth]{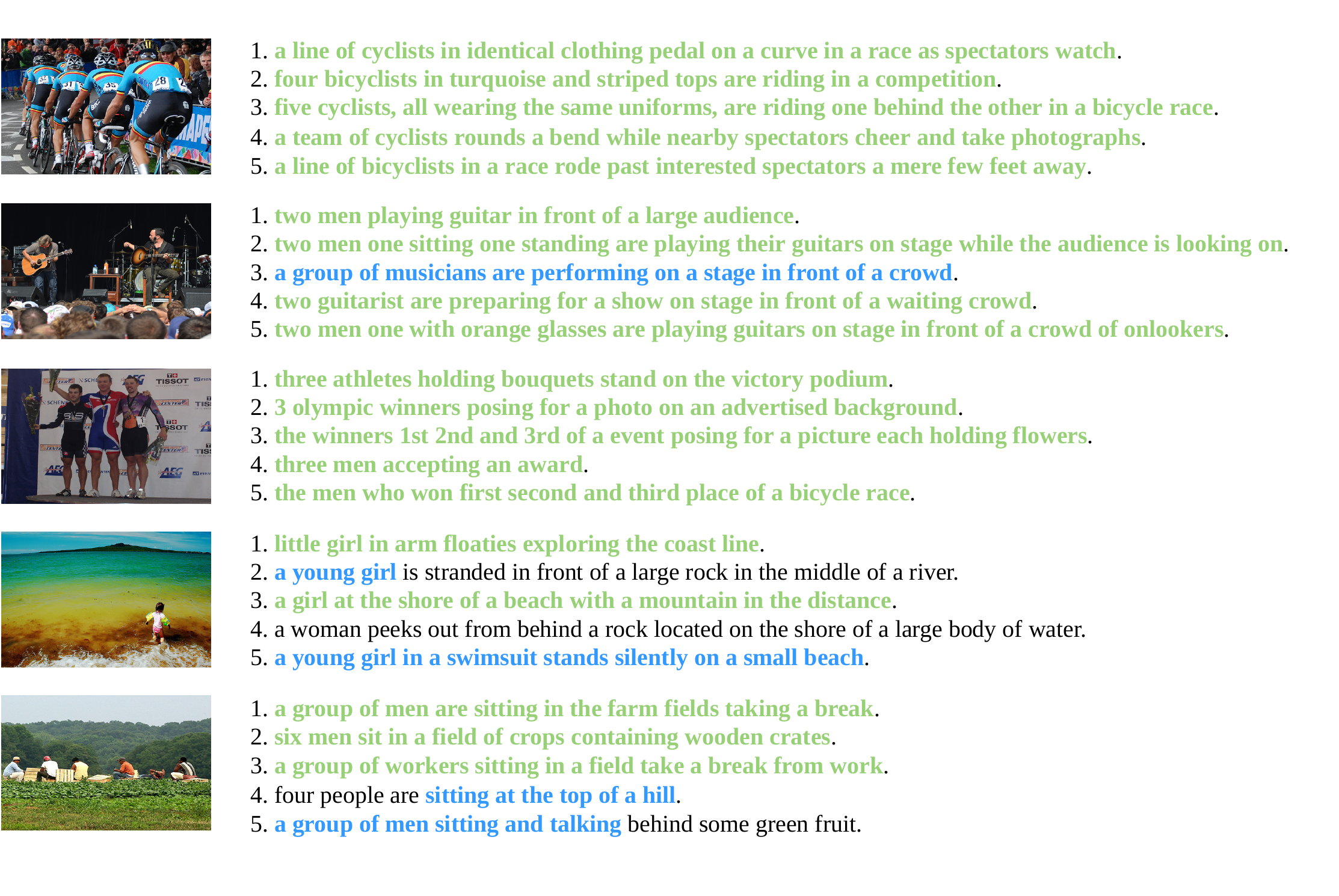}
  \vspace{-12pt}
   \caption{ Examples of image-to-text retrieval results (top 5) of MXM-CLR on Flickr30K\cite{PeterYoung2014FromID} dataset. The green texts are the ground truth captions and the blue texts are manually annotated to indicate the matched content w.r.t the query.
   }
   \label{fig:i2t_flickr}
\end{figure*}

\begin{figure*}[htbp]
  \centering
  \includegraphics[width=0.95\linewidth]{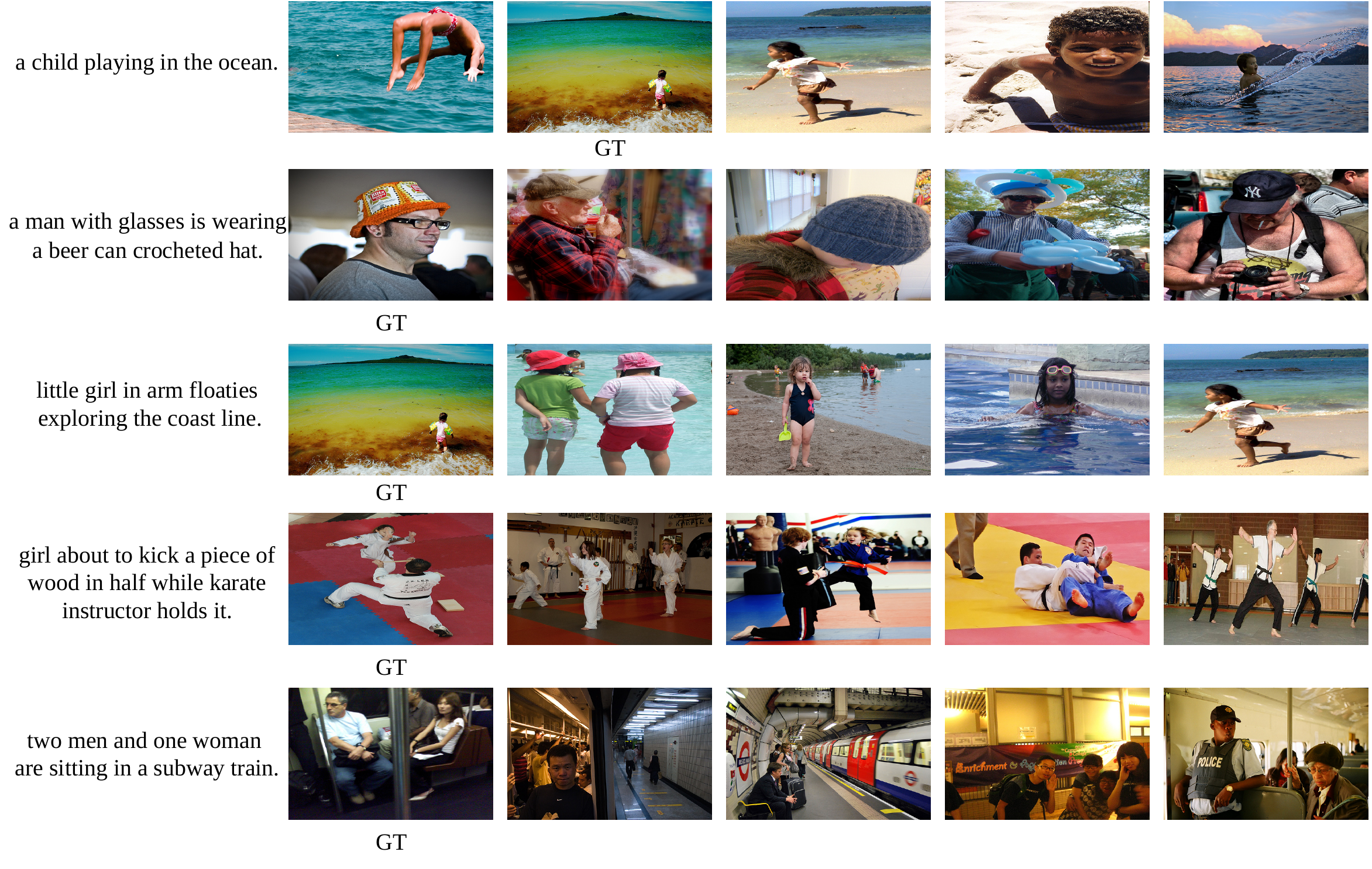}
   \caption{ Examples of text-to-image retrieval results (top 5) of MXM-CLR on Flickr30K\cite{PeterYoung2014FromID} dataset. It can be seen in most cases, the top 1 result returns the ground truth image and other results also consist certain similar content to the query text.
   }
   \label{fig:t2i_flickr}
\end{figure*}

\textbf{More results on Flickr30K dataset.}
Figure \ref{fig:i2t_flickr} and \ref{fig:t2i_flickr} show the qualitative results on the Flicker30K dataset \cite{PeterYoung2014FromID}.
It can be seen MXM-CLR is also generalizable to learn good representations for natural images with complex content.
Meanwhile, it is interesting to see most of the top 5 results returned by MXM-CLR consist similar content instructed by the query, even though they are not annotated as ground truth in the original dataset.
This indicates MXM-CLR is able to learn the semantic features and relations embedded in both modalities.


\end{document}